\documentclass[journal]{IEEEtran}

\usepackage{tikz}
\usepackage{tikz-cd}
\usepackage{float}
\usepackage{comment}
\usepackage{color}
\usepackage{amsmath, amsthm, amssymb}
\newtheorem{problem}{Problem}
\newtheorem{definition}{Definition}
\usepackage[colorlinks=true, linkcolor=black, citecolor=green, urlcolor=blue]{hyperref}
\usepackage{caption} 
\usepackage{enumitem} 
\usepackage{multirow}
\newtheorem{theorem}{Theorem}
\newtheorem{corollary}[theorem]{Corollary}

\begin{document}

\title{Hierarchical GraphCut Phase Unwrapping based on Invariance of
Diffeomorphisms Framework}

\author{
Xiang Gao\textsuperscript{*}, 
Xinmu Wang\textsuperscript{*}, 
Zhou Zhao, 
Junqi Huang, 
Xianfeng David Gu\textsuperscript{\dag}
\vspace{-5mm}
\thanks{
\centerline{\rule{0.4\columnwidth}{0.4pt}}\\[0.5em]
\hspace*{1em}\textsuperscript{*}Authors have contributed equally.\\
\hspace*{1em}\textsuperscript{\dag}Corresponding author: gu@cs.stonybrook.edu. All authors are with the 3D Scanning Lab at the Center of Excellence in Wireless and Information Technology (CEWIT), Stony Brook University, Stony Brook, NY.
}
}

\maketitle

\begin{abstract}
Recent years have witnessed rapid advancements in 3D scanning technologies, with diverse applications spanning VR/AR, digital human creation, and medical imaging. Structured-light scanning with phase-shifting techniques is preferred for its use of non-radiative, low-intensity visible light and high accuracy, making it well suited for human-centric applications such as capturing 4D facial dynamics. A key step in these systems is phase unwrapping, which recovers continuous phase values from measurements that are inherently wrapped modulo \(2\pi\). The goal is to estimate the unwrapped phase count \(k\), an integer-valued variable in the equation \(\Phi = \phi + 2\pi k\), where \(\phi\) is the wrapped phase and \(\Phi\) is the true phase. However, the presence of noise, occlusions, and piecewise continuous phase functions induced by complex 3D surface geometry makes the inverse reconstruction of the true phase extremely challenging. This is because phase unwrapping is an inherently ill-posed problem: measurements only provide modulo \(2\pi\) values, and recovering the correct unwrapped phase count requires strong assumptions about the smoothness or continuity of the underlying 3D surface. Existing methods typically involve a trade-off between speed and accuracy: Fast approaches lack precision, while accurate algorithms are too slow for real-time use. To overcome these limitations, this work proposes a novel phase unwrapping framework that reformulates GraphCut-based unwrapping as a pixel-labeling problem. This framework helps significantly improve the estimation of the unwrapped phase count \(k\) through the invariance property of diffeomorphisms applied in image space via conformal and optimal transport (OT) maps. An odd number of diffeomorphisms are precomputed from the input phase data, and a hierarchical GraphCut algorithm is applied in each corresponding domain. The resulting label maps are fused via majority voting to efficiently and robustly estimate the unwrapped phase count \(k\) at each pixel, using an odd number of votes to break ties. Experimental results demonstrate a 45.5\(\times\) speedup and lower \(L^2\) error in both real experiments and simulations, showing potential for real-time applications.
\end{abstract}

\section{Introduction}
Recent years have seen rapid advancements in 3D scanning technologies, with applications in VR/AR, digital humans, medical imaging, autonomous driving, and robotics. Common approaches include infrared-based systems \cite{infared3Dscanner2012, infared2013}, time-of-flight (ToF) methods \cite{timeofflight2010, timeofflightSuper2008}, and structured-light scanning with phase-shifting techniques \cite{Huang2003Highspeed3S, ZHANG2010149, zhang2018, highresolution2006Song, zhangcalibration2006}. While infrared and ToF methods offer real-time capabilities, they suffer from safety concerns or limited spatial accuracy. In contrast, structured-light systems using visible light patterns provide high-resolution and accurate depth maps, making them ideal for human-centric applications such as 4D facial dynamics. However, their performance heavily depends on robust phase unwrapping to convert wrapped fringes into continuous phase maps for depth and 3D point cloud reconstruction.

This work addresses a core trade-off in phase unwrapping for structured-light 3D scanning: fast methods tend to produce unwrapping artifacts, while accurate ones are often too slow for real-time use. Phase unwrapping is inherently ill-posed and sensitive to noise, occlusion, and surface reflectance. Existing approaches fall into three main categories: path-following \cite{Goldste1988, qualityguided1995}, minimum norm \cite{minNorm1999, multigrid}, and energy-based methods \cite{graphcut2007, kolmogorov2004energy}. Path-following is efficient but prone to error propagation. Minimum norm methods promote smoothness but may oversimplify sharp features. Energy-based methods like GraphCut \cite{graphcut2007} offer high accuracy by minimizing global discontinuity energy, but they are computationally expensive and hard to scale. These challenges point to the need for a method that reduces artifacts while remaining fast and scalable for real-time, high-resolution 3D scanning.

To address these challenges, we reinterpret GraphCut-based phase unwrapping as a pixel-labeling problem, forming the foundation of our novel framework. Operating entirely in image space, we leverage invariance under diffeomorphisms by applying conformal mappings and optimal transport (OT) maps to produce conformally equivalent and area-preserving domains. This enables efficient and accurate application of a hierarchical GraphCut algorithm to obtain the unwrapped phase count through majority voting, which is necessary for computing depth and reconstructing the 3D point cloud. Our framework yields substantial gains in computational efficiency and scalability. Experiments show a \textbf{45.5× speedup} and the \textbf{lowest $L^2$ error} among all compared methods. An ablation study further supports its effectiveness in real-world settings, especially near abrupt surface changes where it helps avoid unwrapping artifacts.
\noindent To summarize, we offer \textbf{three principal contributions}: \begin{itemize}[noitemsep, topsep=0pt, leftmargin=*] \item To the best of our knowledge, this is the first work to formulate one of the fundamental properties of low-level vision operators: \textbf{Invariance under Diffeomorphisms (ID)}. \item We propose a novel phase unwrapping framework to exploit the invariance property, the ID framework, by combining conformal mappings, optimal transport (OT) maps, and \textbf{ID Hierarchical GraphCut phase unwrapping}. \item Extensive real-world experiments and simulations demonstrate the effectiveness and robustness of our method, achieving a \textbf{45.5× speedup} and the \textbf{lowest $L^2$ error} compared to state-of-the-art approaches.  \end{itemize}

\vspace{2mm}

\section{Related Works}
\noindent{\textbf{Traditional Phase Unwrapping.}} Traditional phase unwrapping methods are broadly categorized into path-following, minimum \(L^p\)-norm, and minimum discontinuity approaches. Path-following methods, such as Goldstein’s algorithm \cite{Goldste1988}, use branch cuts to balance residues, while quality-guided methods \cite{qualityguided1995} improve efficiency by directing unwrapping paths with quality maps, though they remain sensitive to noise. Reliability phase unwrapping \cite{wenjing2004realiabilityreview} further refines this approach by employing enhanced quality maps. Minimum \(L^p\)-norm methods, like Poisson solvers for robustness against noise through global phase continuity optimization but suffer from slow convergence. Minimum discontinuity methods, such as Flynn’s algorithm \cite{flynn1996consistent}, segment the wrapped phase into regions and assign integer multiples of \(2\pi\) to minimize discontinuities, excelling where path-following methods fail. Despite their strengths, these traditional approaches struggle to balance computational efficiency with robustness.
 
\noindent{\textbf{Energy Based Phase Unwrapping. }}
The use of GraphCut-based methods \cite{graphcut2007} significantly advanced phase unwrapping by framing it as an energy minimization problem, enabling higher quality unwrapping results. In many fundamental computer vision tasks, such as image restoration, stereo matching, segmentation, and phase unwrapping, the goal is to assign labels (e.g., intensity, disparity, segmentation regions, or phase counts) to pixels based on noisy measurements. With the presence of uncertainties, the optimal labeling can be determined by minimizing an energy function that reflects the problem's constraints and objectives. For example, minimum cut/maximum flow algorithms \cite{kolmogorov2004energy, boykov2004experimental, lombaert2005multilevel, veksler2008star} are widely used for energy minimization, where a graph is constructed such that the minimum cut on the graph corresponds to the optimal energy configuration. The classical use of energy minimization is in solving the pixel labeling problem, where the variables represent individual pixels, and the possible values represent attributes such as displacements or intensities. By leveraging this energy-based framework, GraphCut provides a more accurate and globally consistent phase unwrapping solution, particularly in challenging scenarios with noise and other measurement uncertainties.\\

\noindent \textbf{Pixel Labeling Problem.} Suppose \( \Omega \subset \mathbb{R}^2 \) is a planar domain, and a signal \( I: \Omega \to \mathbb{R} \) is defined over \( \Omega \). In practice, \( \Omega \) is discretized into a set of pixels \( \mathcal{P} \), and the signal values are polluted by noise. Thus, the observed signal is a noisy image \( \hat{I}: \mathcal{P} \to \mathbb{R} \).
\begin{problem}[\textbf{Pixel Labeling}] The input is a set of pixels $\mathcal{P}$ with a noisy signal $\hat{I}$ and a set of labels $\mathcal{L}$ . The goal is to find a labeling $f^*:\mathcal{P}\to\mathcal{L}$ that minimizes a special energy $E(f)$,
\[
    f^* := \underset{f:\mathcal{P} \to \mathcal{L}}{\text{argmin}} \, E(\hat{I}, f),
\]
such that the optimal label \( f^* \) and the noisy signal \( \hat{I} \) can be used to recover the authentic signal \( I \).
\end{problem}
\noindent The energy function \( E(f) \) is chosen depending on properties of the signal on the labelled regions, such as piecewise consistency in the neighborhoods or discontinuities at boundaries. There are many research works for solving the pixel labeling problem based on energy optimization techniques. However, due to the complexity of real situations and the variety of noises, finding a robust way to solve the pixel labeling problem still remains a challenge. 
\vspace{-3mm}
\section{Theoretical Foundation}
\noindent \textbf{Invariance under Diffeomorphisms.}  Our key idea comes from a natural property of the pixel labeling problem, the continuity or discontinuity of the signal $I$ is invariant under the diffeomorphic deformations. Suppose $\mathcal{G}$ is a subgroup of the diffeomorphism group of $\Omega$, each element $g\in\mathcal{G}$ is a smooth invertible mapping $g:\Omega\to\Omega$. By dense discretizations, we can treat $g$ as an automorphism of the pixel set $\mathcal{P}$ to itself, $g:\mathcal{P}\to\mathcal{P}$. For each mapping $g\in\mathcal{G}$, it deforms the image $\hat{I}$ to $g(\hat{I})$, and the best label of $g(\hat{I})$ becomes $f_g$. Then the labeling is invariant under $g$: for each pixel $p\in\mathcal{P}$ with the original label $f(p)$ is mapped to $g(p)$, with the new label $f_g(g(p))$, then $f(p)=f_g(g(p))$. Namely, the following diagram commutes:

\begin{center}
\begin{tikzcd}[column sep=small]
\mathcal{P} \arrow{rr}{g} \arrow[swap]{dr}{f}& &\mathcal{P} \arrow{dl}{f_g}\\
& \mathcal{L} & 
\end{tikzcd}    
\end{center}
for some type of image $\hat{I}:\mathcal{P}\to\mathbb{R}$ and any diffeomorphism $g\in \mathcal{G}$.
In the following, we define the property of invariance under diffeomorphism. All the images form the image space (functional space) denoted as $\mathcal{I}$. An image processing algorithm is treated as an operator $\mathcal{T}:\mathcal{I}\to\mathcal{I}$, which maps each image $I\in \mathcal{I}$ to another image. 

\begin{figure*}[t]
    \centering
    \begin{tabular}{cccc}
    \includegraphics[width=0.225\textwidth]{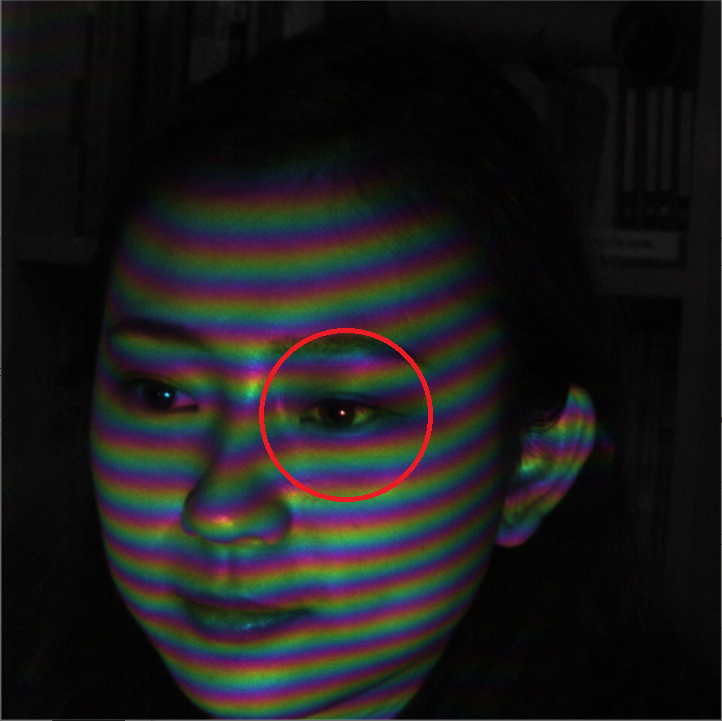}&
    \includegraphics[width=0.225\textwidth]{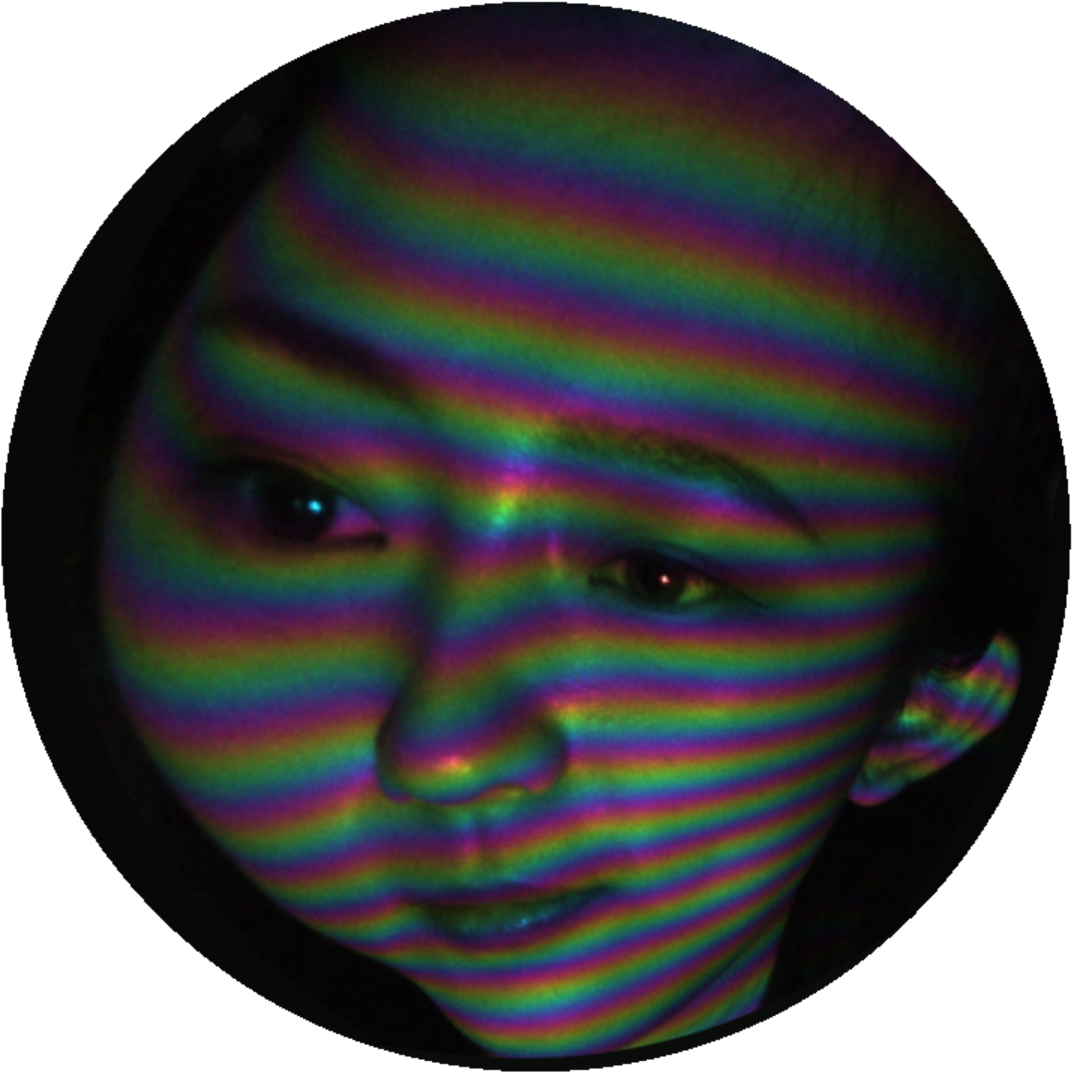}&
    \includegraphics[width=0.225\textwidth]{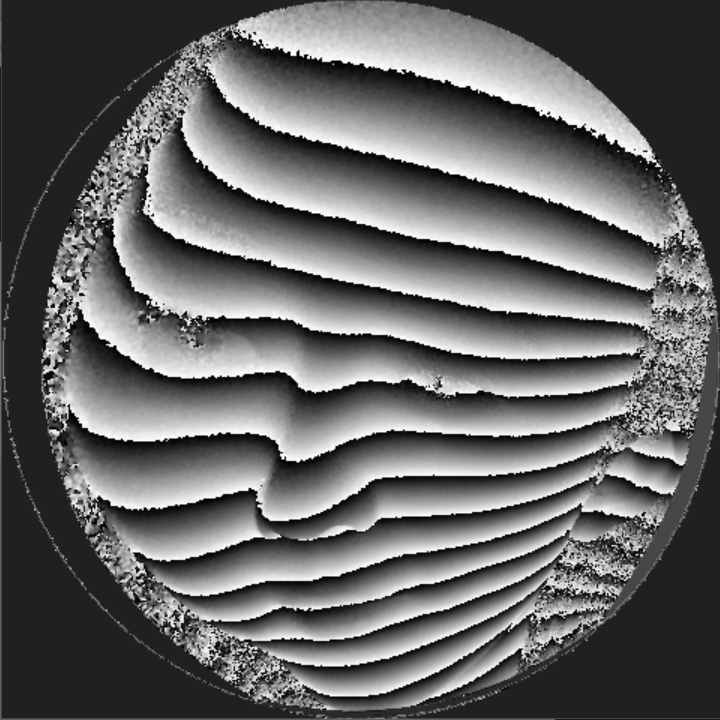}&
    \includegraphics[width=0.225\textwidth]{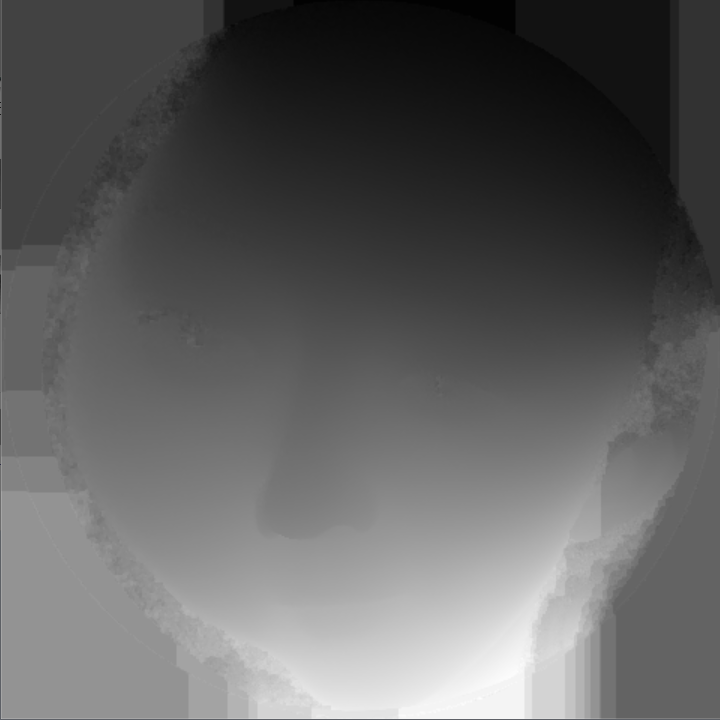}\\
    
    \includegraphics[width=0.225\textwidth]{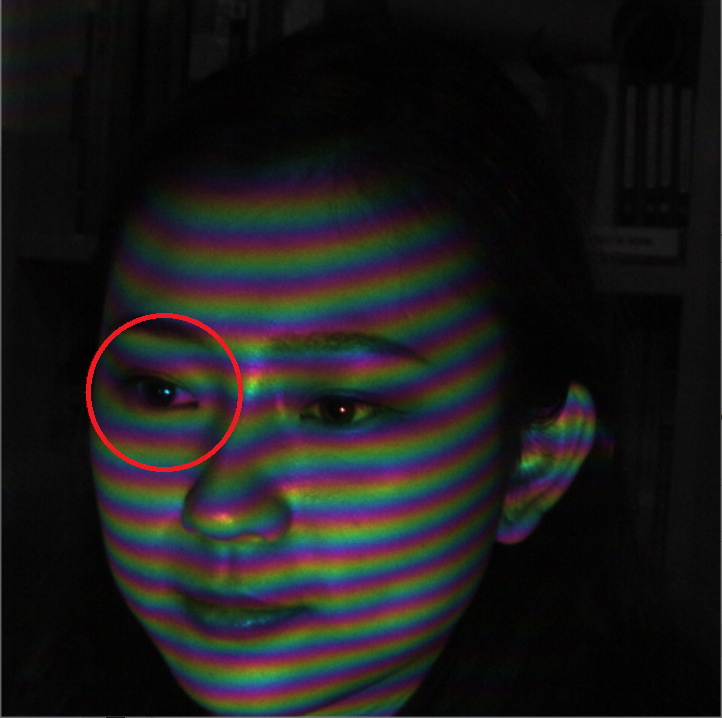}&
    \includegraphics[width=0.225\textwidth]{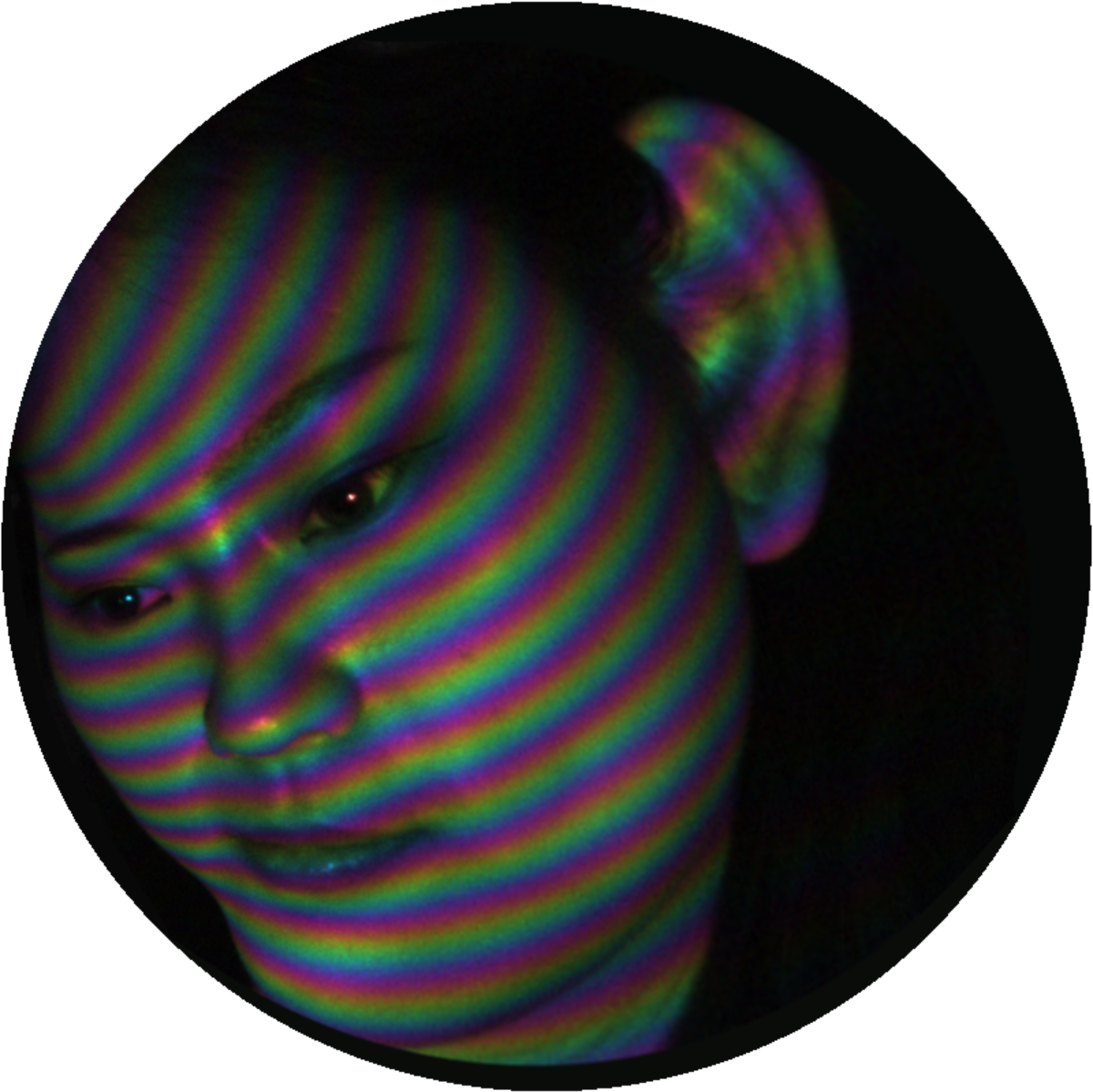}&
    \includegraphics[width=0.225\textwidth]{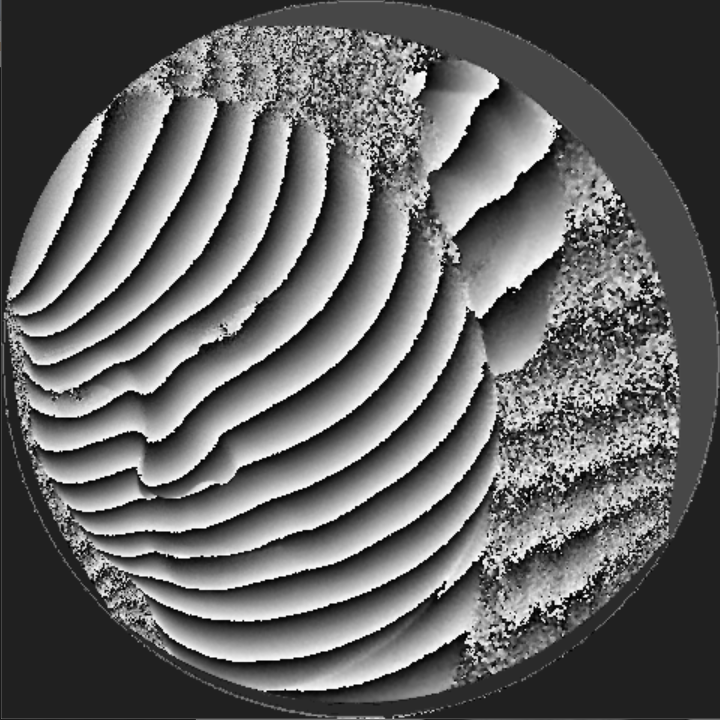}&
    \includegraphics[width=0.225\textwidth]{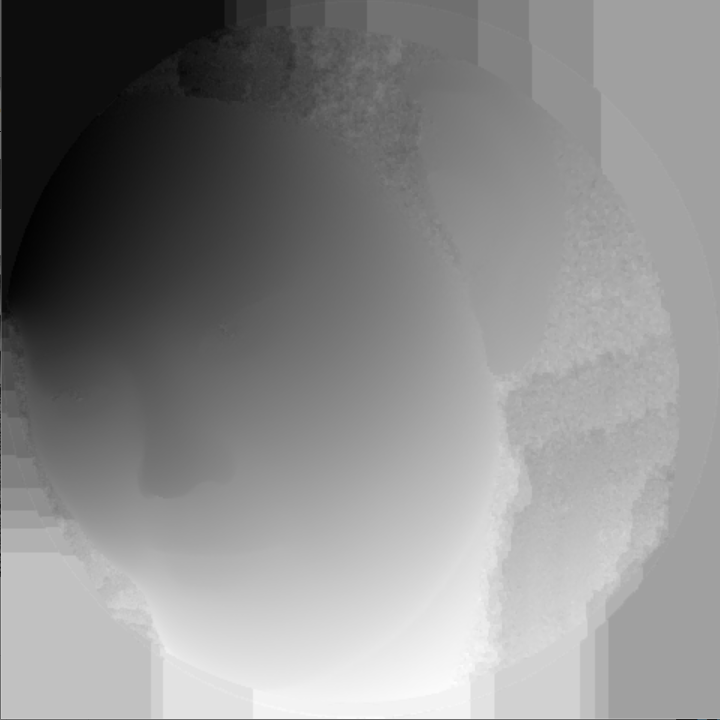}\\
    
    \includegraphics[width=0.225\textwidth]{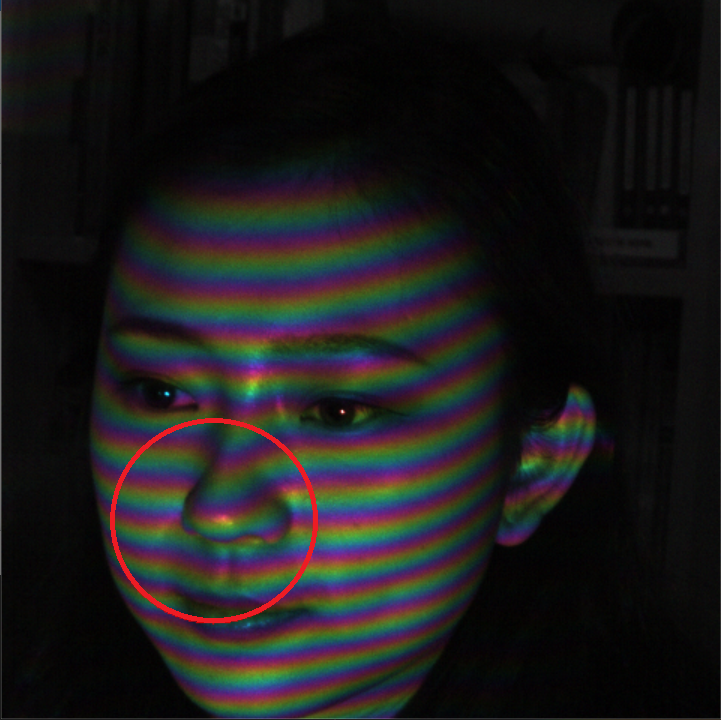}&
    \includegraphics[width=0.225\textwidth]{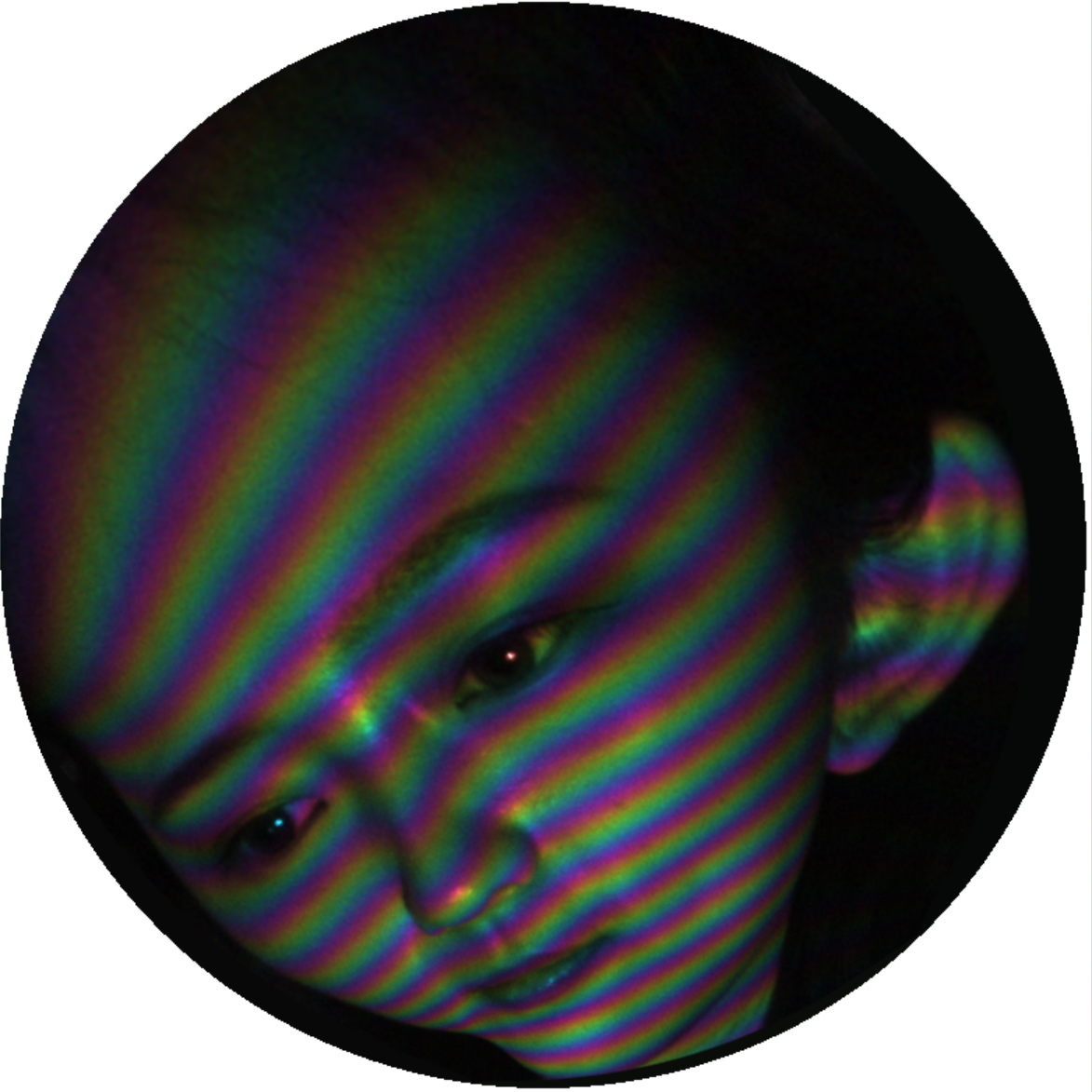}&
    \includegraphics[width=0.225\textwidth]{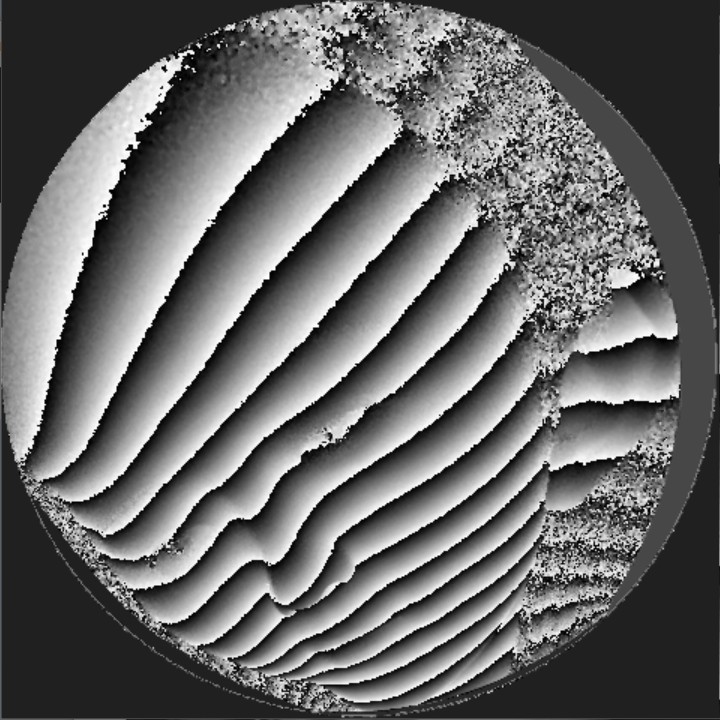}&
    \includegraphics[width=0.225\textwidth]{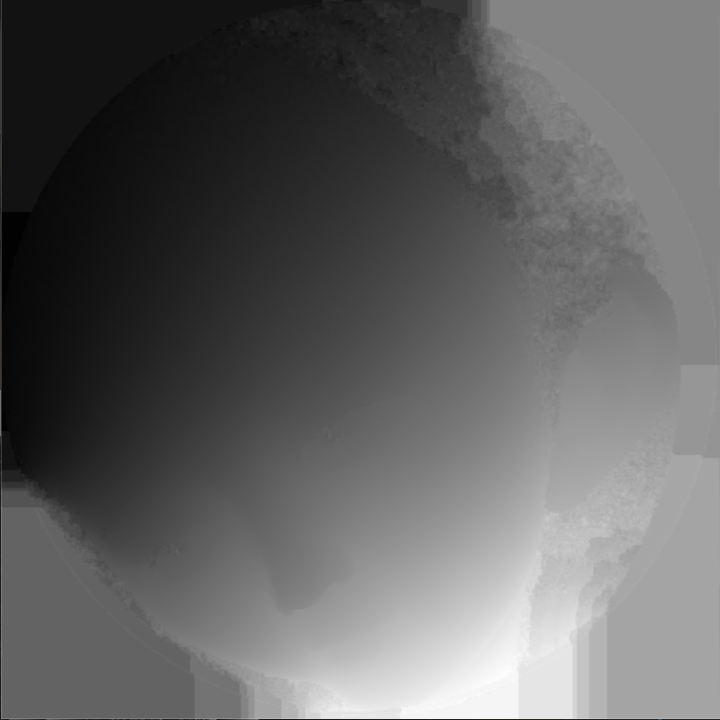}\\
    
    (a). Fringe Patterns (\( I_{0,1,2}\))  &(b). \textbf{Conformal Maps} (\( \overset{\sim}{I}_{0,1,2} \)) &(c). Wrapped Phases (\( \varphi_{0,1,2}\)) &(d). Unwrapped Phases (\( \Phi_{0,1,2}\))\\

    \end{tabular}
    \caption{
    Fringe patterns (\( I_{0,1,2} \)) are deformed via \textbf{M\"obius Transformations} in Eq.~\ref{eqn:Mobius_transformation} to produce conformal maps (\( \overset{\sim}{I}_{0,1,2} \)). These maps are then used to extract the unwrapped phases (\( \varPhi_{0,1,2} \)) via \textbf{ID Hierarchical GraphCut Algorithm}.
\label{fig:LV_ID_CF}} 
\vspace{-5mm}
\end{figure*} 

\begin{definition}[Invariance under Diffeomorphism] Suppose the operator $\mathcal{T}:\mathcal{I}\to\mathcal{I}$ satisfies the following property: for any image $I\in \mathcal{I}$, any two diffeomorphisms $g_1,g_2\in \mathcal{G}$, and any pixel $p\in\mathcal{P}$, we have 
\begin{equation}
    \mathcal{T}[g_1(I)](p) = \mathcal{T}[g_2(I)](p),
\end{equation}
then we say the operator $\mathcal{T}$ is invariant under the diffeomorphism group $\mathcal{G}$.
\end{definition}

\noindent \textbf{Phase Unwrapping Problem.} In computer vision, fringe projection is a widely used method for 3D shape acquisition, where sinusoidal stripe patterns are projected onto a surface to determine its geometry. Using at least three patterns recorded in sequence, the absolute phase $\Phi:\mathcal{P}\to\mathbb{R}$, which encodes projector coordinates, can be recovered. Each pixel $p$ in the camera image corresponds to a point $r$ on the surface, linked via a ray passing through projector pixel $q$, and the depth of $r$ is computed using triangulation. Practically, only the relative phase $\varphi$, obtained as $\Phi(p) = \varphi(p) + 2k(p)\pi$, is measurable, where $k:\mathcal{P}\to\mathbb{Z}$ is the \emph{wrap count}, and recovering $k$ is known as \emph{phase unwrapping}. This problem involves minimizing the $L^2$ norm of the gradient of $\Phi$, representing its smoothness, approximated in discrete form using finite differences. A graph $G$ is constructed, where pixels are nodes, adjacent pixels form edges, and source/sink nodes connect to all pixels. Edge capacities, based on the discrete harmonic energy of $\Phi$, enable iterative binary optimizations using max flow/min cut algorithms, solving for $k$ efficiently.

\noindent \textbf{Invariance under Diffeomorphism Framework.} In the following, we show the phase unwrapping problem has the property of invariance under diffeomorphism.\\
\vspace{-5mm}
\begin{corollary}[\textbf{ID property of Phase Unwrapping}]
Suppose the geometric surface to be captured is smooth, then the phase unwrapping operator is invariant under the conformal diffeomorphism group.
\end{corollary}

\renewcommand{\qedsymbol}{} 
\begin{proof}
Since the geometric surface to be captured is smooth, the absolute phase as a function defined on the image domain $\Phi:\Omega\to \mathbb{R}$ is also smooth. The energy used to optimize the phase count function is the measurement of the smoothness of $\Phi$, generally an approximation of the harmonic energy of $\Phi$. According to the main theorem~\ref{thm:main}, the harmonic energy is invariant under conformal mappings. Therefore, the optimal phase count function is invariant under conformal mappings. $\square$
\end{proof} \unskip
Similarly, optimal transport (OT) maps are diffeomorphisms and preserve the continuities and the sharp features, therefore can be applied for the same purpose as well.

\begin{figure*}[htbp]
    \centering
    \begin{tabular}{cccc}
        \includegraphics[width=0.225\textwidth]{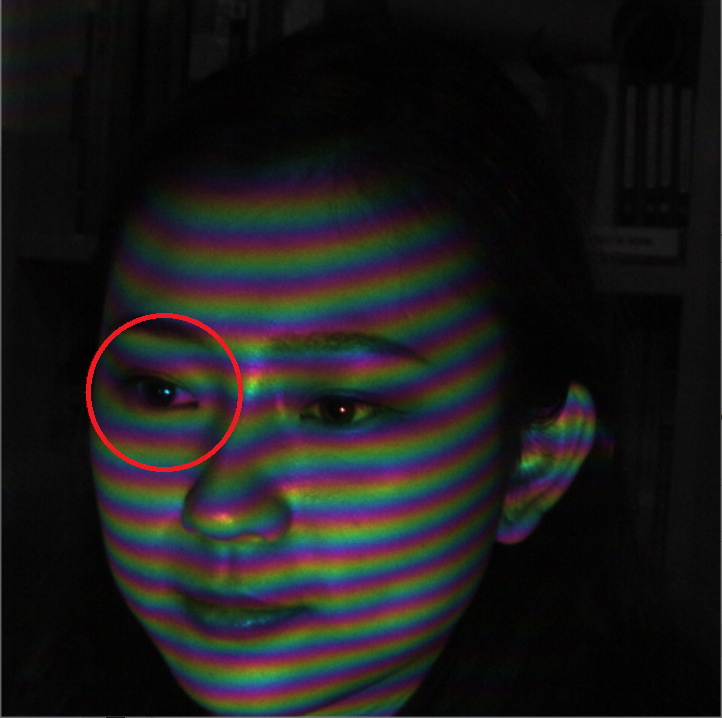} &
        \includegraphics[width=0.225\textwidth]{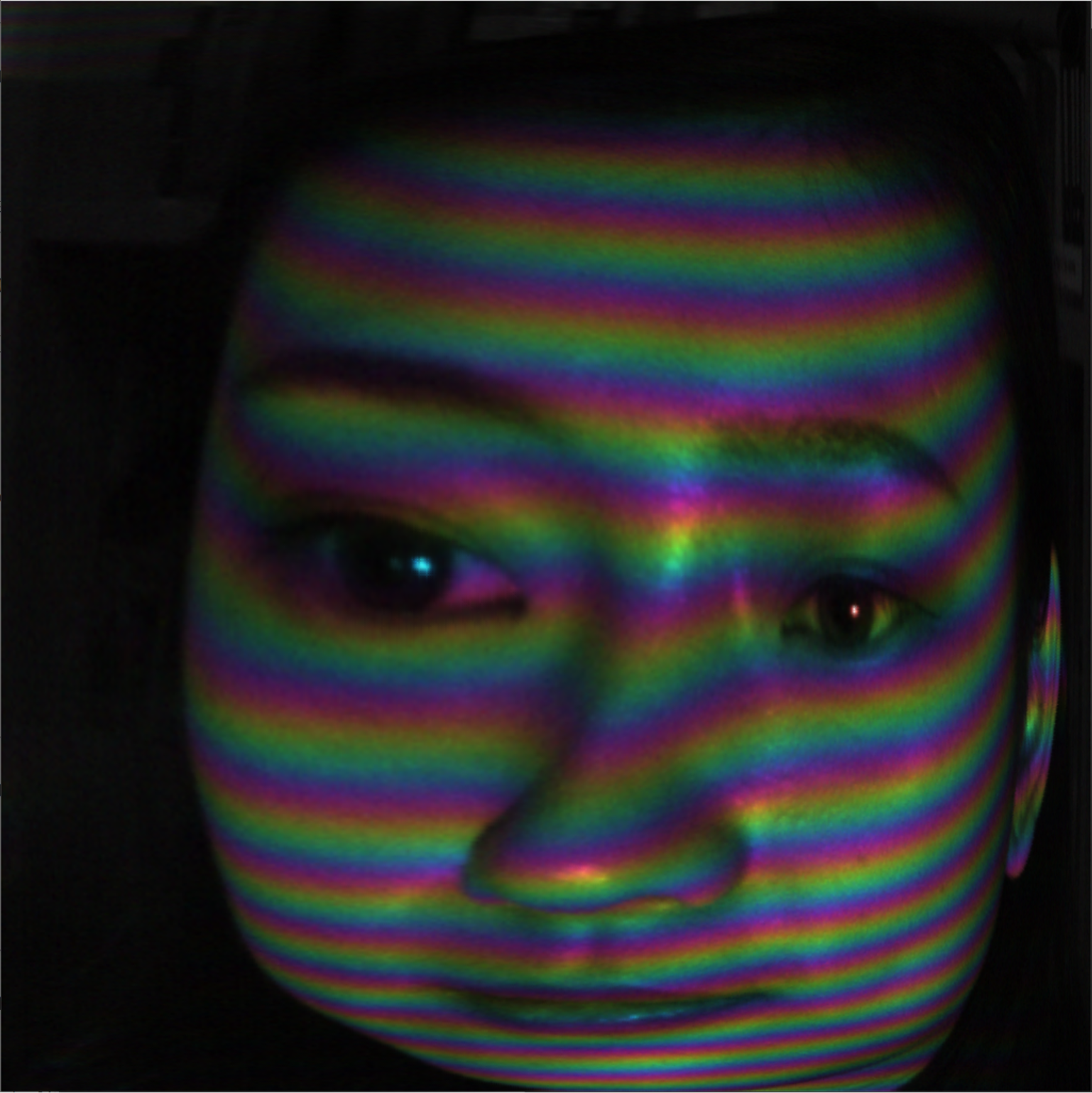} &
        \includegraphics[width=0.225\textwidth]{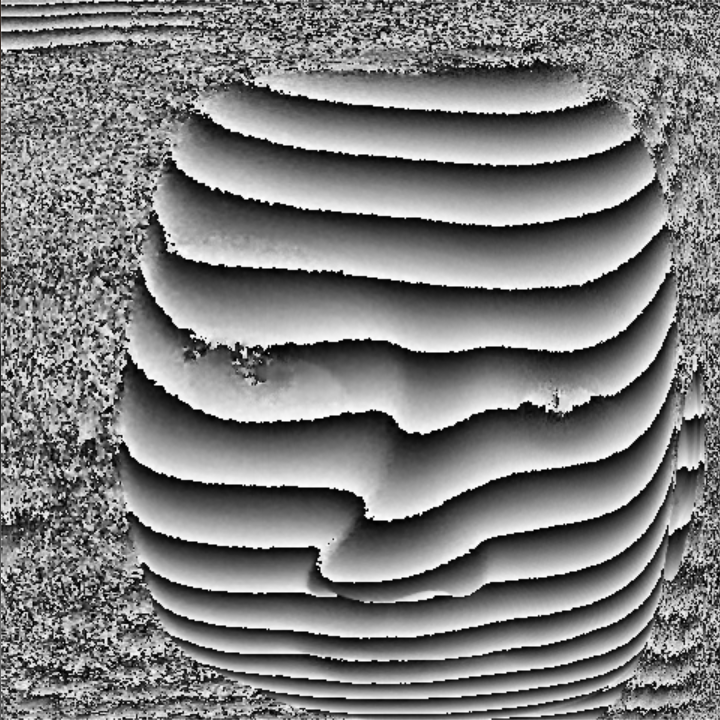} &
        \includegraphics[width=0.225\textwidth]{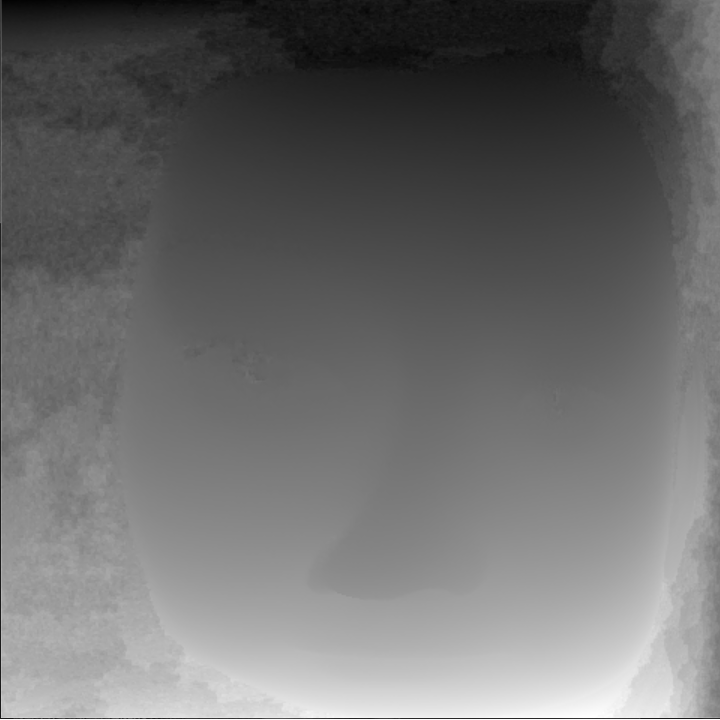} \\
        
        \includegraphics[width=0.225\textwidth]{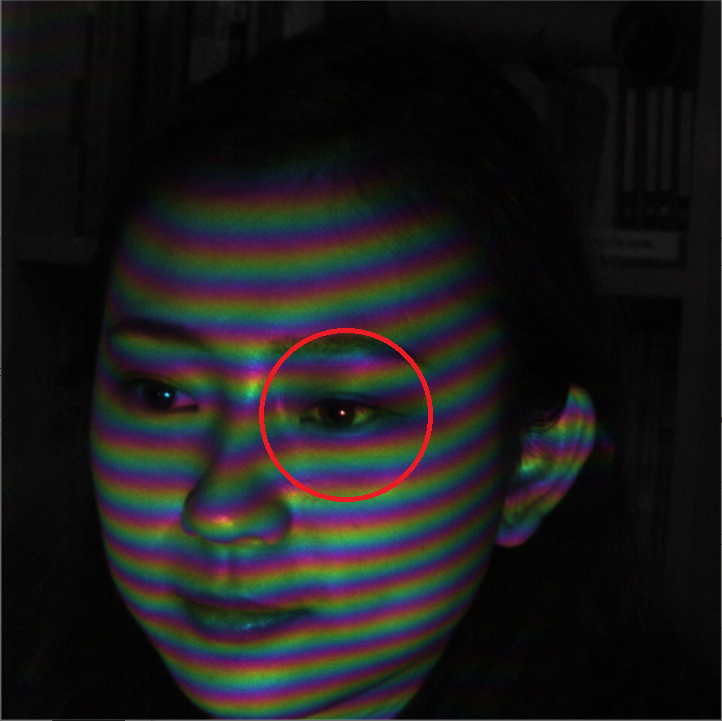} &
        \includegraphics[width=0.225\textwidth]{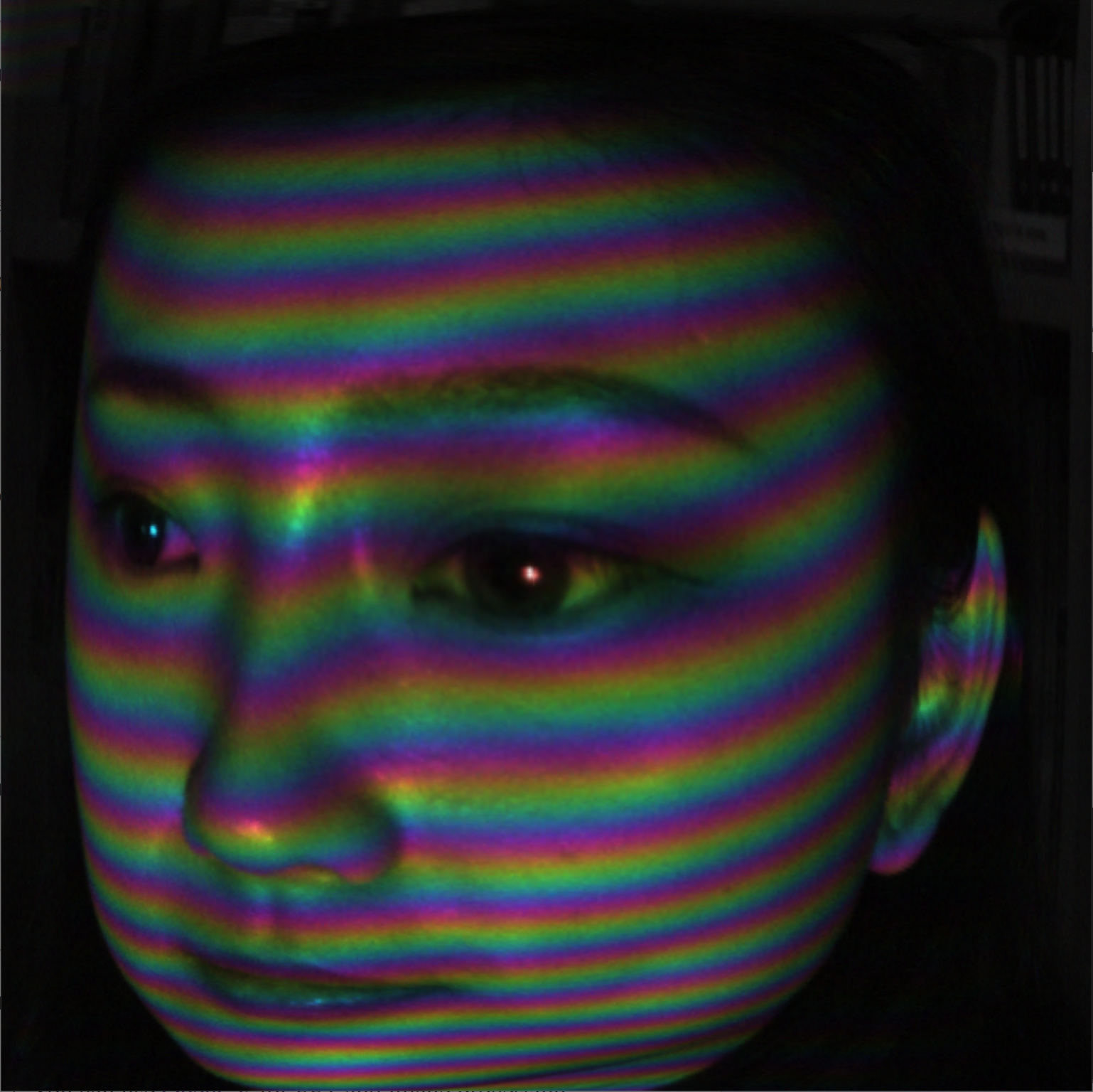} &
        \includegraphics[width=0.225\textwidth]{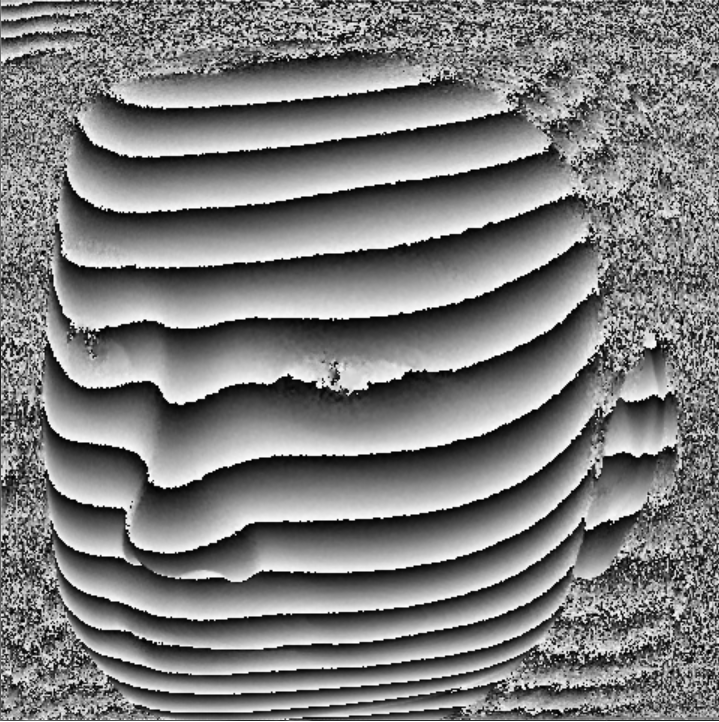} &
        \includegraphics[width=0.225\textwidth]{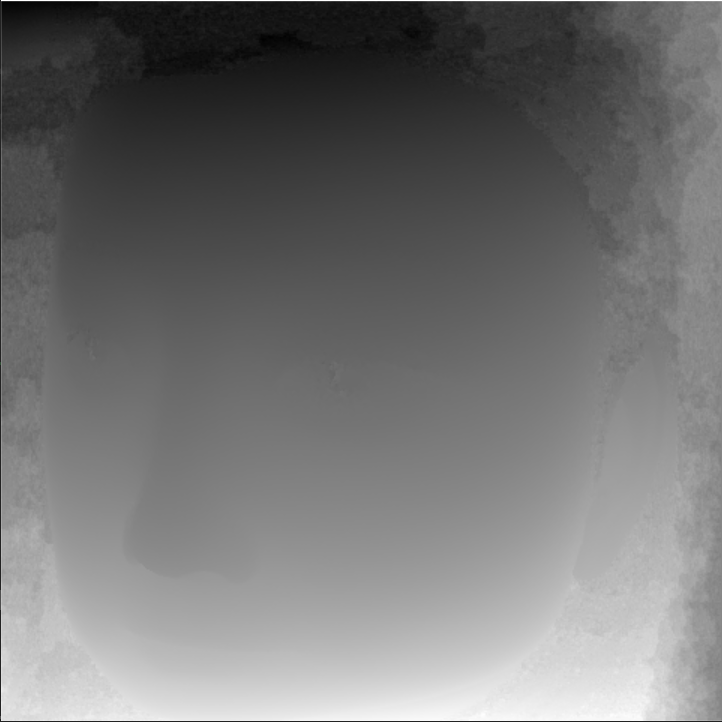} \\
        
        \includegraphics[width=0.225\textwidth]{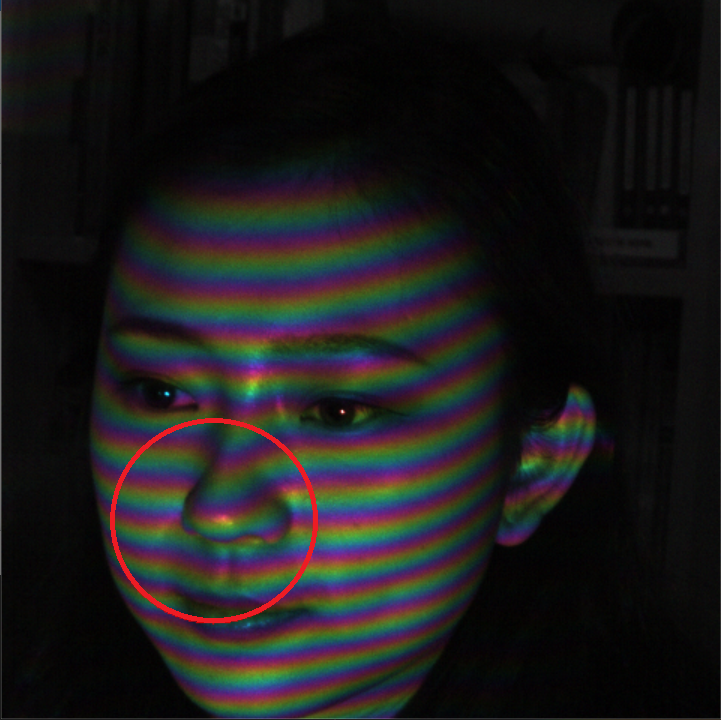} &
        \includegraphics[width=0.225\textwidth]{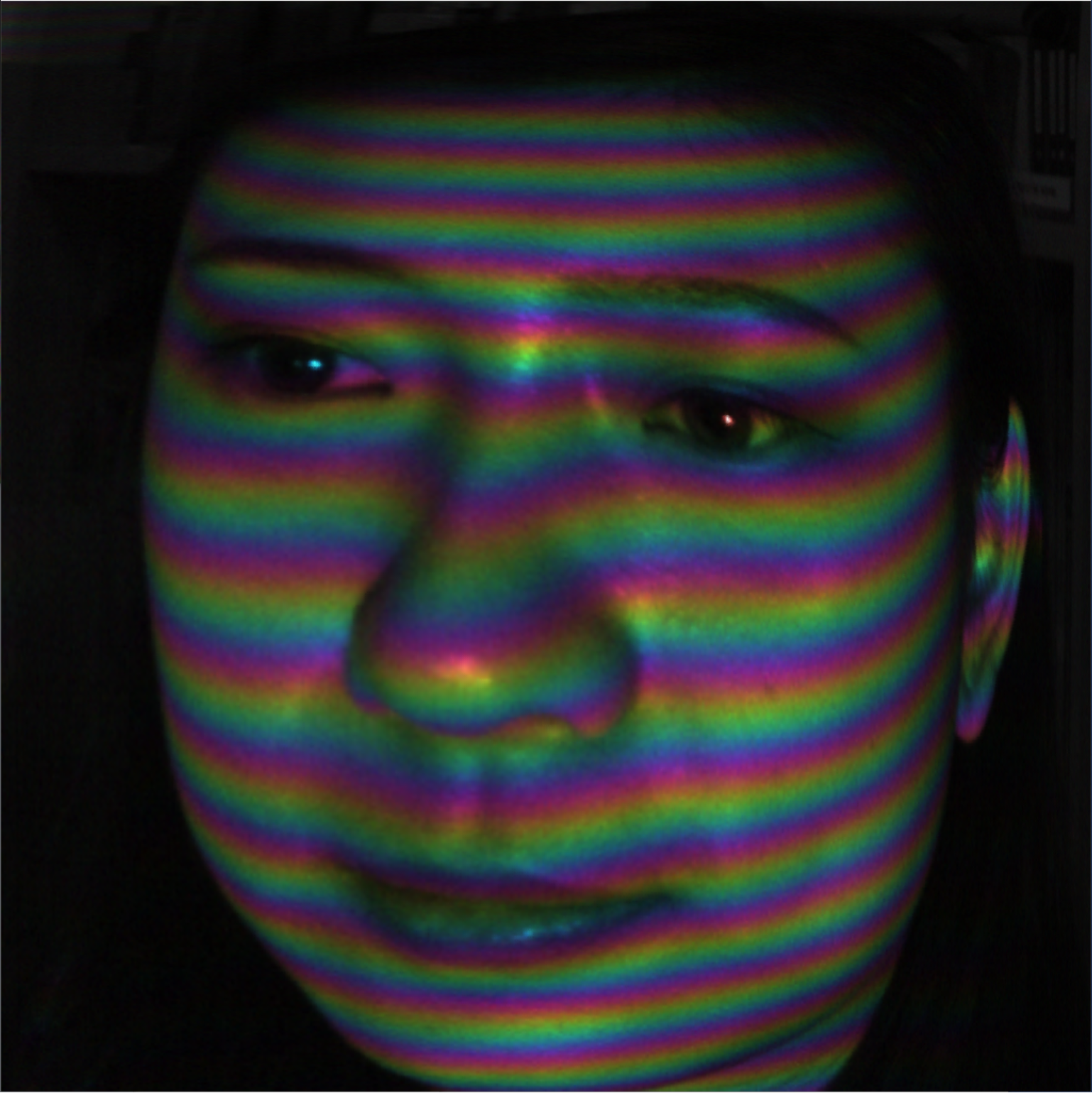} &
        \includegraphics[width=0.225\textwidth]{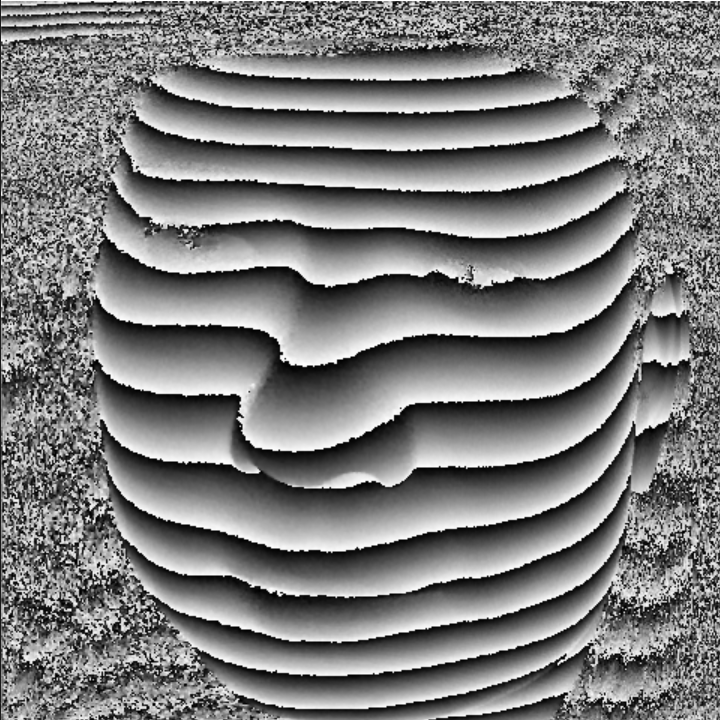} &
        \includegraphics[width=0.225\textwidth]{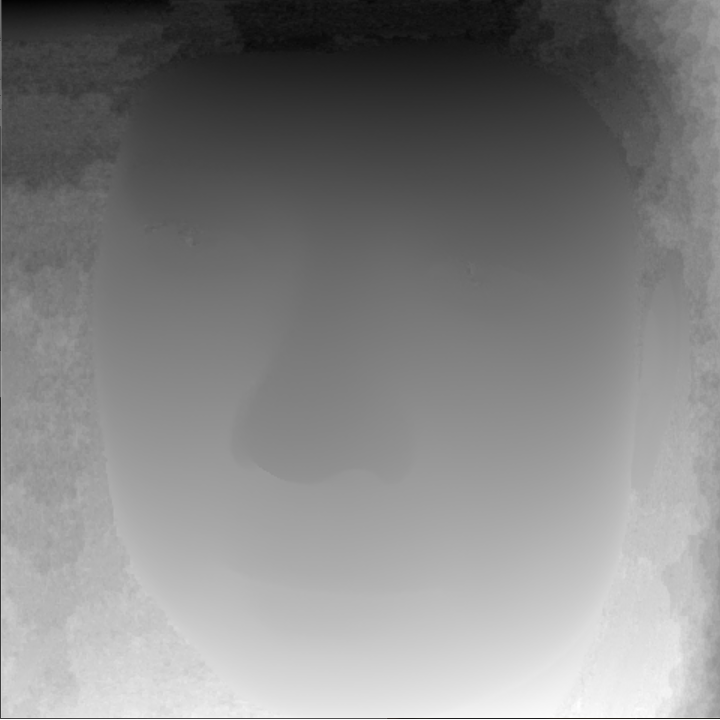} \\
        
        (a) Fringe Patterns (\( I_{0,1,2} \)) & (b) \textbf{OT Maps (\( \overset{\sim}{I}_{0,1,2} \))} & (c) Wrapped Phases (\( \varphi_{0,1,2} \)) & (d) Unwrapped Phases (\( \Phi_{0,1,2} \)) \\
    \end{tabular}
    \caption{Fringe patterns (\( I_0, I_1, I_2 \)) are deformed using diffeomorphic Optimal Transport maps computed from \textbf{\( N = 3 \) ROIs}, which correspond to the circled regions in red. The resulting maps (\( \overset{\sim}{I}_0, \overset{\sim}{I}_1, \overset{\sim}{I}_2 \)) yield wrapped phases (\( \varphi_0, \varphi_1, \varphi_2 \)), which are unwrapped using the \textbf{ID Hierarchical GraphCut} to obtain \( \Phi_0, \Phi_1, \Phi_2 \).}
    \label{fig:LV_ID_OT}

    \vspace{5mm} 

    \begin{tabular}{cccc}
        \includegraphics[width=0.225\textwidth]{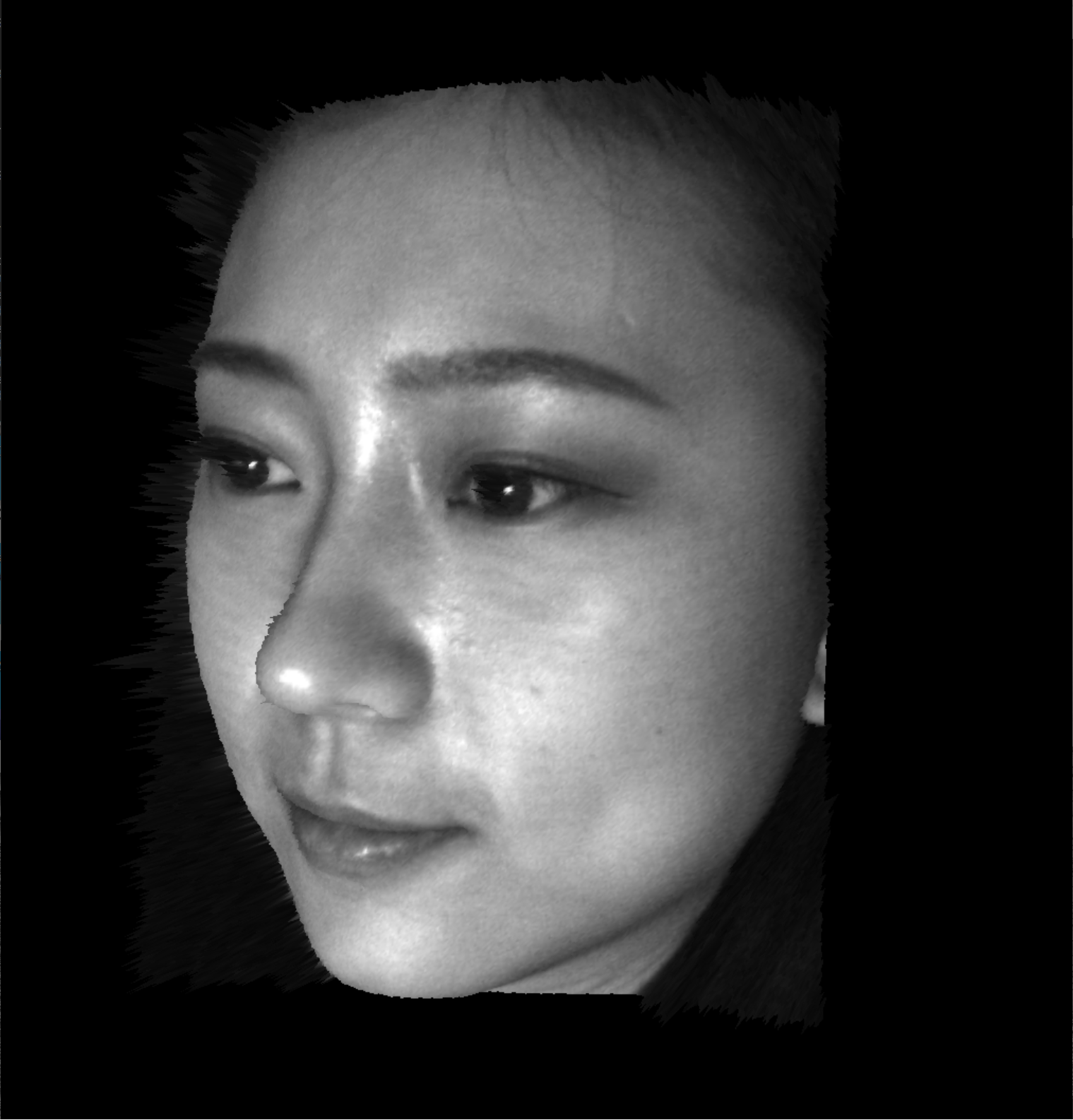} &
        \includegraphics[width=0.225\textwidth]{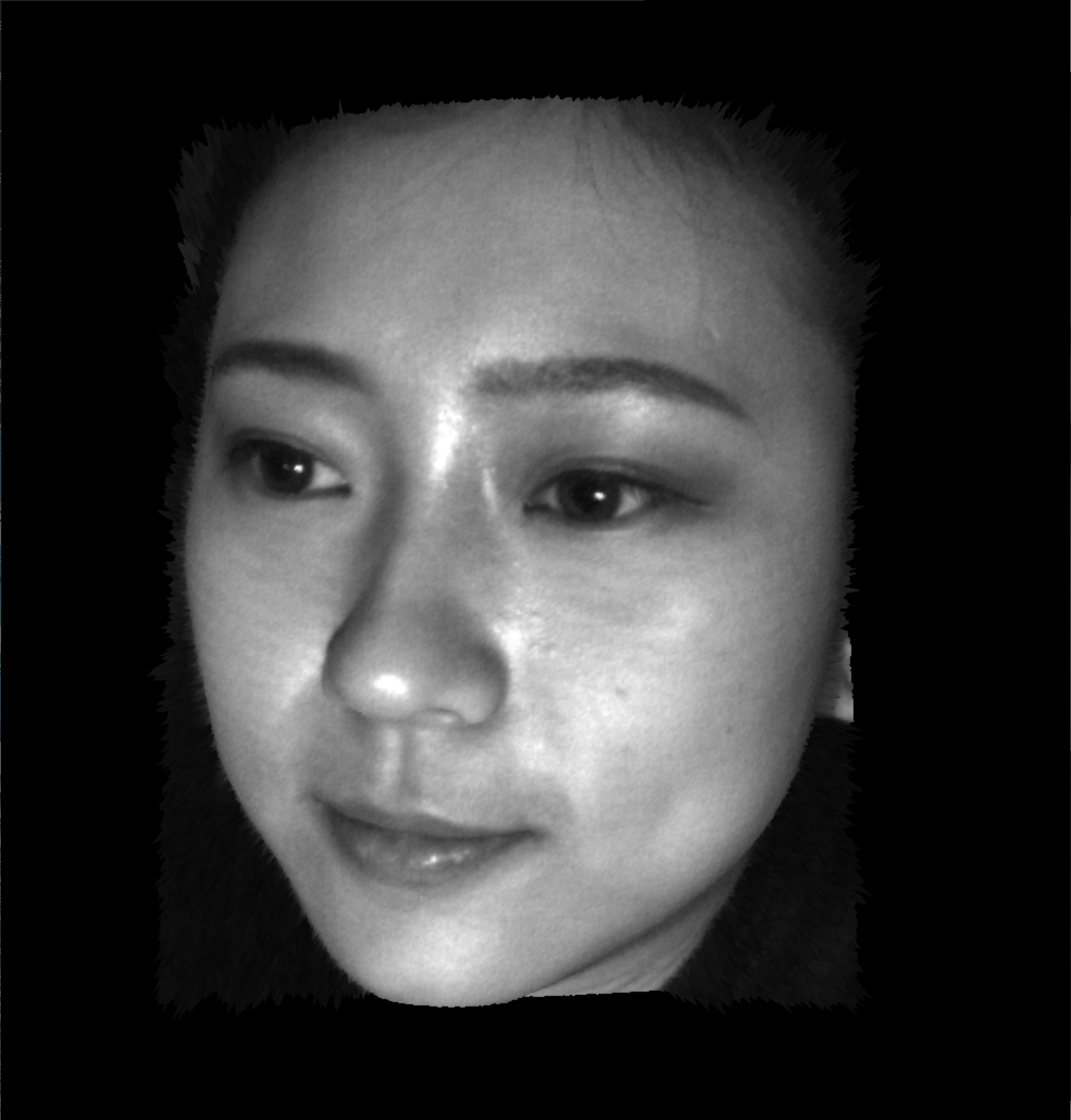} &
        \includegraphics[width=0.225\textwidth]{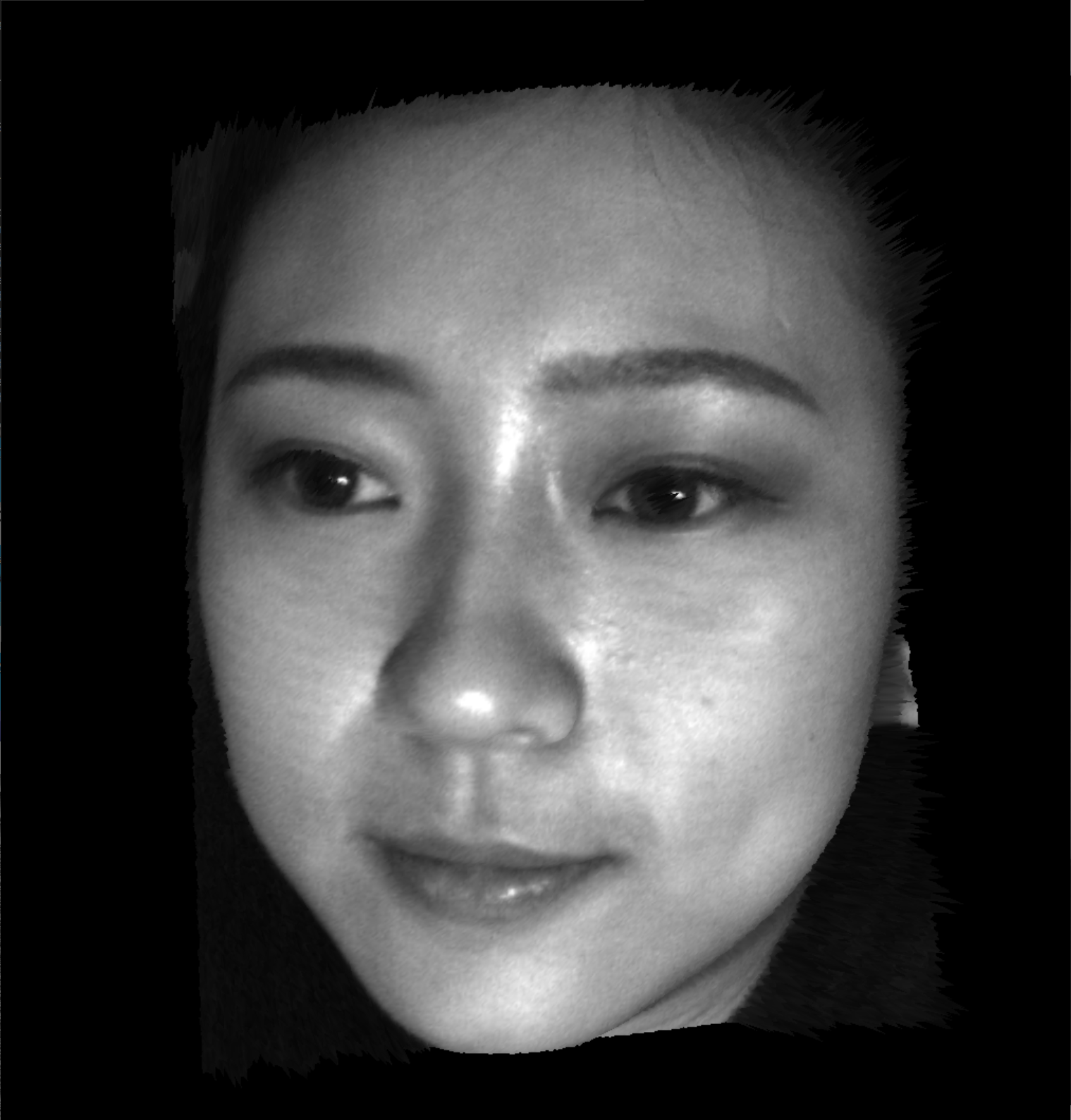} &
        \includegraphics[width=0.225\textwidth]{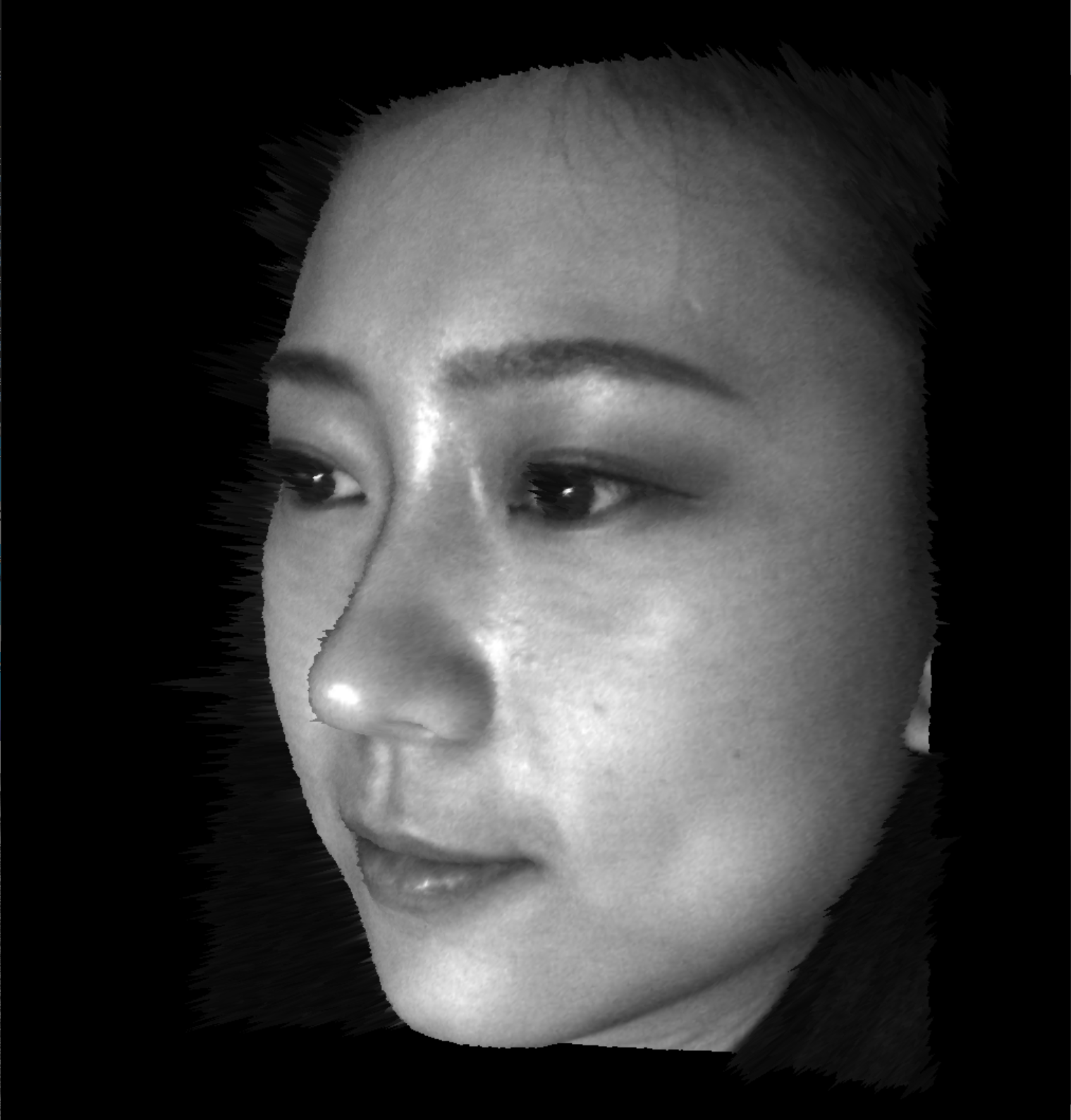} \\
    \end{tabular}
    \caption{Reconstructed 3D surface (Point Cloud) viewed from different camera angles based on the unwrapped phase recovered by applying \textbf{ID Hierarchical GraphCut} on the deformed wrapped phases in \textbf{Fig.~\ref{fig:LV_ID_CF}} and \textbf{Fig.~\ref{fig:LV_ID_OT}}.}
    \label{fig:LV_Point_cloud}
\end{figure*}

In our current work, we propose to use two special types of diffeomorphisms. One is the conformal map (CM) and the other is the OT map, both of which are diffeomorphic. We choose conformal maps because the harmonic energy of a signal is invariant under conformal mappings, and the harmonic energy is one of the most popular ways to measure the smoothness of the function. We choose OT maps because they can fully control the target area measure, so that we can zoom in on regions of interests to improve the unwrapping accuracy. Basically, we can choose several regions of interests $\Omega_k$, $k=1,2,\dots,n$, and increase the densities on $\Omega_k$, then by solving a Monge-Amp\`ere equation to obtain the OT maps $g_k$'s, where $g_k$ enlarges $\Omega_k$ for better analysis as shown in \textbf{Fig.~\ref{fig:LV_ID_OT}}. Similarly, we can choose several M\"obius transformations $g_k$'s using the closed form Eqn.~(\ref{eqn:Mobius_transformation}), as shown in \textbf{Fig.~\ref{fig:LV_ID_CF}}. The deformed image $g_k(\mathcal{I})$ is fed into the phase unwrapping algorithm based on hierarchical GraphCut to obtain a phase count function $\mathbf{k_i}$. Thus, for each pixel $p\in\mathcal{P}$, we obtain $n$ phase counts \( \{\mathbf{k}_1 \circ g_1^{-1}(p),\ \mathbf{k}_2 \circ g_2^{-1}(p),\ \ldots,\ \mathbf{k}_n \circ g_n^{-1}(p)\} \), the final phase count for $p$ can be determined by the major voting scheme among $\mathbf{k_i}\circ g_i^{-1}$'s. In order to avoid a tie, we choose $n$ to be an odd number. We call this approach the \textbf{Invariance under Diffeomorphism (ID)} framework for phase unwrapping.

\noindent \textbf{This work} heavily depends on the construction of diffeomorphisms between planar domains. Here, we briefly introduce the concepts and theorems of optimal transport maps and conformal maps. We refer the reader to~\cite{villani2003optimal} and~\cite{CCG_2007} for thorough treatments. 

\begin{definition}[\textbf{Diffeomorphism}] 
Suppose $\Omega,\Omega^*\subset\mathbb{R}^2$ are planar domains (simply connected open sets). A differentiable mapping $T:\Omega\to\Omega^*$ is called a diffeomorphism if it is bijective (both surjective and injective), and its inverse $T^{-1}:\Omega^*\to \Omega$ is also differentiable. If these mappings are $k$ times continuously differentiable, $T$ is called a $C^k$-diffeomorphism.
\end{definition}

\noindent $T$ is diffeomorphic if it is proper and its differential $DT_x:\mathbb{R}^2\to\mathbb{R}^2$ is bijective, and hence a linear isomorphism; namely, the Jacobian matrix $DT(x)$ is non-degenerate everywhere, at each point $x\in \Omega$. 

\noindent{\textbf{Optimal Transport Map.}} Given two probability distributions $f(x)dx$ and $g(y)dy$ with supports $\Omega$ and $\Omega^*$ respectively. A $C^1$ map $T:\Omega\to\Omega^*$ is \emph{measure preserving} if $\text{det}DT(x) = f(x)/g\circ T(x)$, where $DT$ is the \emph{Jacobian} of the map, and denoted as $T_\# f=g$. Given a \emph{cost} function $c:\Omega\times\Omega^*\to\mathbb{R}$ representing the cost for transporting a unit mass from point $x$ to $y$, the \emph{transport cost} of the mapping $T$ is defined as $\mathcal{C}(T) := \int_\Omega c(x,T(x))f(x)dx$. Monge raised the problem of finding the \emph{optimal transport map}, which is the measure-preserving map with the least transportation cost,
\[
    \min \{\mathcal{C}(T): T:\Omega\to\Omega^*,T_\#f = g\}.
\]
Brenier's theorem gives an answer to Monge's problem, under mild regularity conditions:
\begin{theorem}[\textbf{Brenier}] 
If the transport cost is the quadratic Euclidean distance, $c(x,y) = |x-y|^2/2$, then there exists a convex function $u:\Omega\to\mathbb{R}$, unique up to a constant, such that the gradient map $T=D u: \Omega\to\Omega^*$ is the unique optimal transport map. $u$ is called the Brenier potential, satisfying the \emph{Monge-Amp\`ere equation}:
\begin{equation}
    \text{det}~T(x)=\text{det}~D^2u(x) = \frac{f(x)}{g\circ D u(x)},
    \label{eqn:Monge_Ampere}
\end{equation}
with the boundary condition: $T(\Omega)=D u(\Omega)=\Omega^*$.
\label{thm:Brenier}
\end{theorem}
\noindent The regularity of the Brenier potential $u$ (equivalently the optimal transport map $T=Du$) is guaranteed by Caffarelli's theory below.
\begin{theorem}[\textbf{Caffarelli}] Suppose $\Omega$ and $\Omega^*$ are bounded domains in $\mathbb{R}^n$, $\bar{\Omega}$ is the closure of $\Omega$, $C^{2,\alpha}(\Omega)$ is the Hölder space,
and $f$ and $g$ are mass densities on $\Omega$ and $\Omega^*$ satisfying, 
\begin{enumerate}
    \item if $f,g>0$ and $f\in C^\alpha(\Omega)$ and $g\in C^\alpha(\Omega^*)$, and $\Omega^*$ is convex, then $u\in C^{2,\alpha}(\Omega)$;
    \item if $f,g>0$ and $f\in C^\alpha(\Omega)$ and $g\in C^\alpha(\Omega^*)$, and both $\Omega$ and $\Omega^*$ are uniformly convex and $\partial\Omega$ and $\partial\Omega^*$ are $C^{2,\alpha}$, then $u\in C^{2,\alpha}(\bar{\Omega})$.
\end{enumerate}
\end{theorem}

\noindent In practice, we set the source density to be the uniform distribution (Lebesgue measure), and the target density to be the Gaussian mixture with $C^\infty$ regularity, the existence and the uniqueness of optimal transport map is guaranteed by Brenier's theorem, the smoothness of the map by Caffarelli's theorem.  \\
\noindent{\textbf{Conformal Map.}} Suppose $w:\Omega\to\mathbb{C}$ is a complex function, which can be treated as a planar map. We define the complex differential operators, $\partial_z = (\partial_x - i\partial_y)/2$ and  $\partial_{\bar{z}}=(\partial_x + i\partial_y)/2$.
\begin{definition}[\textbf{Conformal Map}]
A complex function $w:\Omega\to\mathbb{C}$ is holomorphic, if $\partial_{\bar{z}} w = 0$ everywhere. If $w$ is invertible, and the inverse is also holomorphic, then $w$ is biholomorphic, namely a planar conformal map. 
\end{definition}
\begin{definition}[\textbf{Harmonic Energy}]
Suppose $f:\Omega\to\mathbb{R}$ is a real-valued function, the harmonic energy of $f$ is 
\[
    E(f):=\int_\Omega |\nabla f|^2 dx \wedge dy = i\int_\Omega (|w_z|^2 + |w_{\bar{z}}|^2) dz\wedge d\bar{z}.
\]
\end{definition}
\begin{theorem}[\textbf{Main}]
\label{thm:main}
Suppose $f:\Omega\to\mathbb{R}$ is a real-valued function, $w:\Omega\to\mathbb{C}$ is a conformal map, then the harmonic energy is invariant under the conformal mapping $w$.
\end{theorem}

\begin{definition}[\textbf{M\"obius Transformation}] A conformal mapping $\varphi:\mathbb{D}^2\to \mathbb{D}^2$ from the unit disk $\mathbb{D}^2=\{z\in\mathbb{C}|~|z|<1\}$ to itself is a M\"obius transformation, which has the form:
\begin{equation}
    w(z)= e^{i\theta} \frac{z-z_0}{1-\bar{z}_0z}, \quad z(w) = e^{-i\theta}\frac{w+e^{i\theta}z_0}{1+\overline{e^{i\theta}z_0}w}\quad 
    \label{eqn:Mobius_transformation}
\end{equation} where $\theta\in [0,2\pi),~|z_0|<1.$
\end{definition}

\noindent Hence, any simply connected planar domain can be conformally mapped to the unit disk by the following Riemann mapping theorem.
\begin{theorem}[\textbf{Riemann Mapping}]
If $\Omega$ is a non-empty simply connected open subset of the complex plane $\mathbb{C}$ which is not all of $\mathbb{C}$, then there exists a biholomorphic mapping $f$ from $\Omega$ onto the open unit disk $\mathbb{D}^2$. All such kinds of mappings differ by a M\"obius transformation.
\end{theorem}
\noindent In practice, we can use Riemann mapping to map the image to the unit disk, and compose it with a M\"obius transformation. This kind of mapping preserves the harmonic energy of the signal.
\vspace{-3mm}
\section{Computational Algorithms}
\begin{figure*}[htbp]
    \centering
    \begin{tabular}{cccc}
        \includegraphics[width=0.23\textwidth]{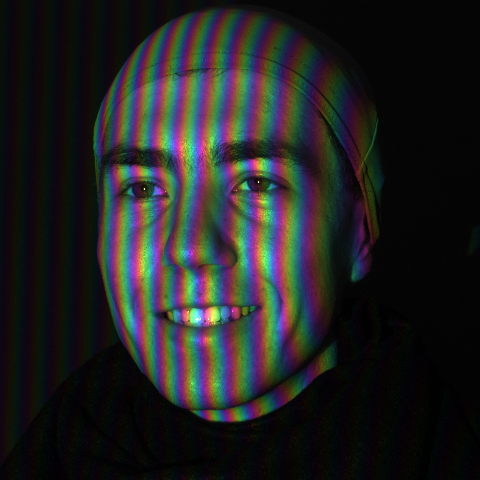} &
        \includegraphics[width=0.23\textwidth]{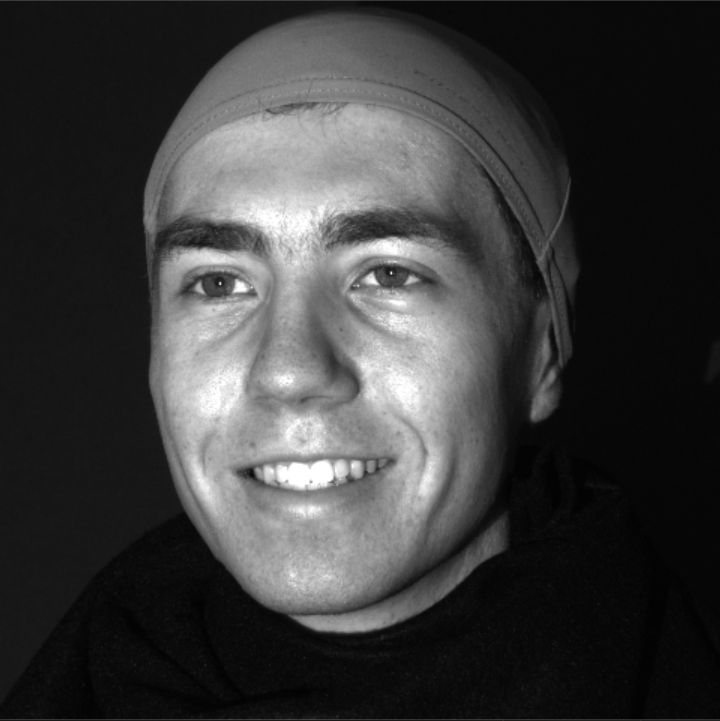} &
        \includegraphics[width=0.23\textwidth]{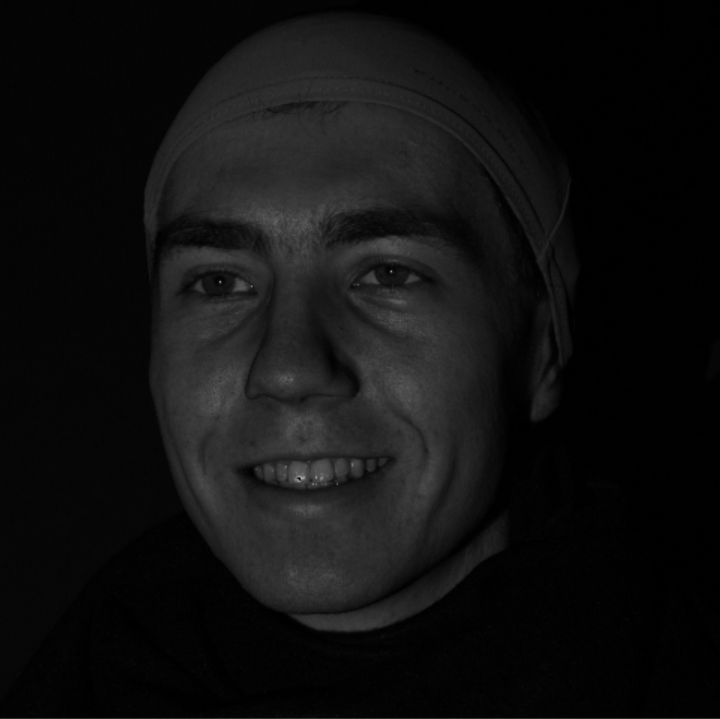} &
        \includegraphics[width=0.23\textwidth]{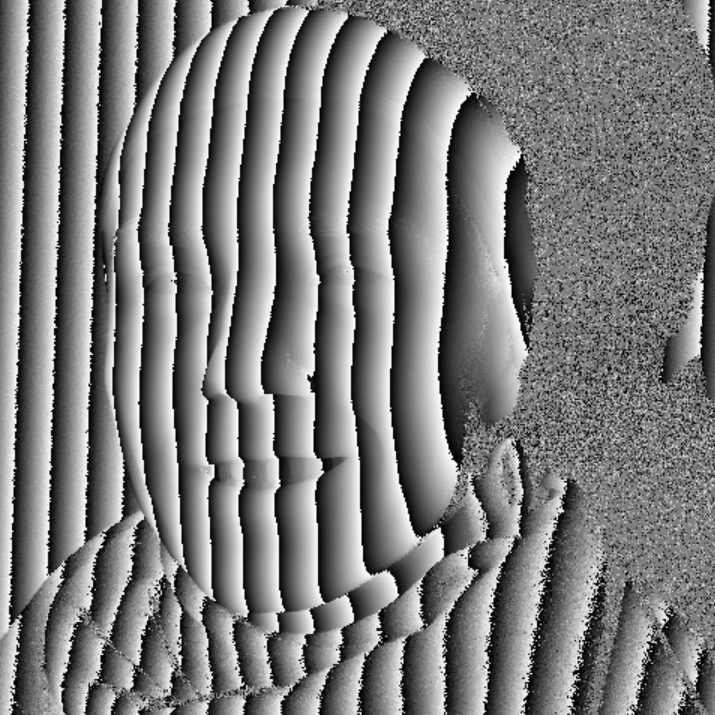} \\
        (a). Fringe Pattern (\( I\)) & (b). Ambient \( A(p) \) & (c). Modulation \( R(p) \) & (d). Wrapped phase \( \varphi(p) \) \\
    \end{tabular}
    \caption{Phase Shifting Pipeline: From left to right, we show the captured fringe images (packed into RGB channels), followed by the computed \textbf{ambient term} \( A(p) \), the \textbf{modulation term} \( R(p) \), and the resulting \textbf{wrapped phase} \( \varphi(p) \).}
    \label{fig:Andrew_phase_shifting}

    \vspace{5mm} 

    \begin{tabular}{ccc}
        \includegraphics[width=0.32\textwidth]{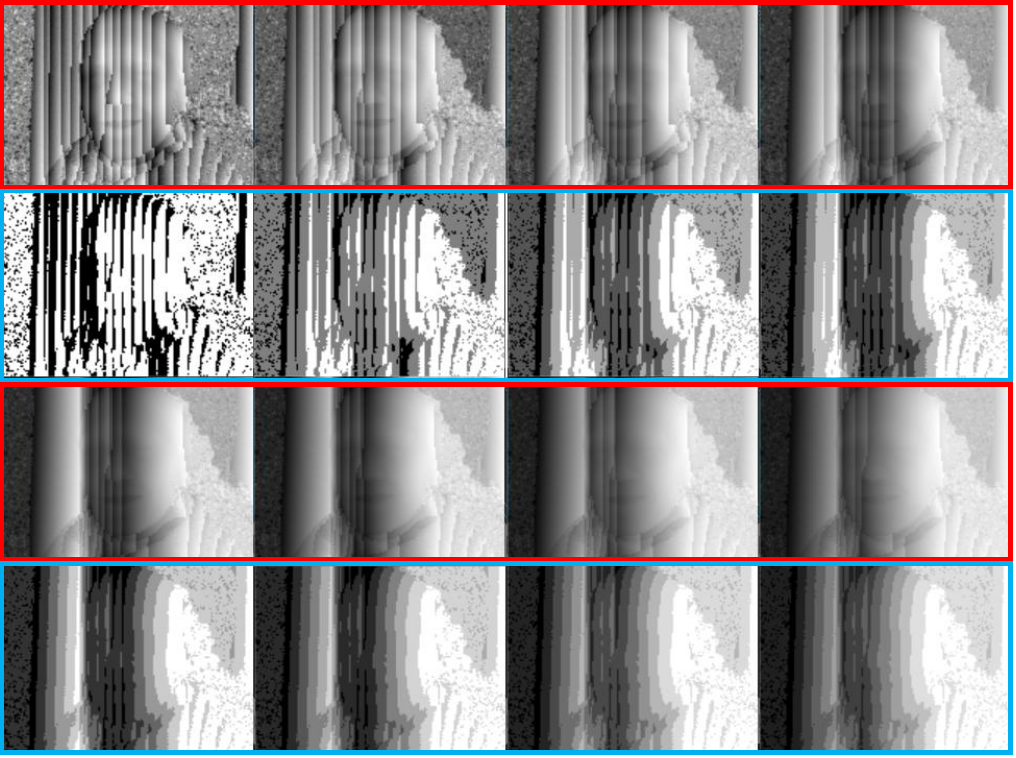} &
        \includegraphics[width=0.32\textwidth]{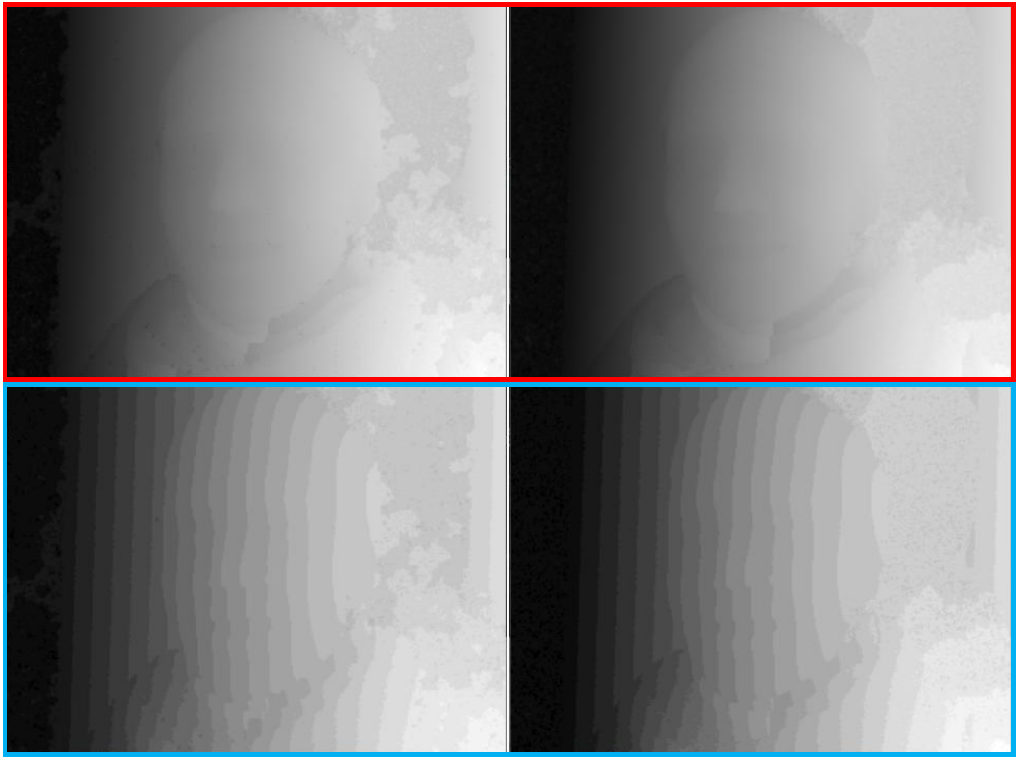} &
        \includegraphics[width=0.32\textwidth]{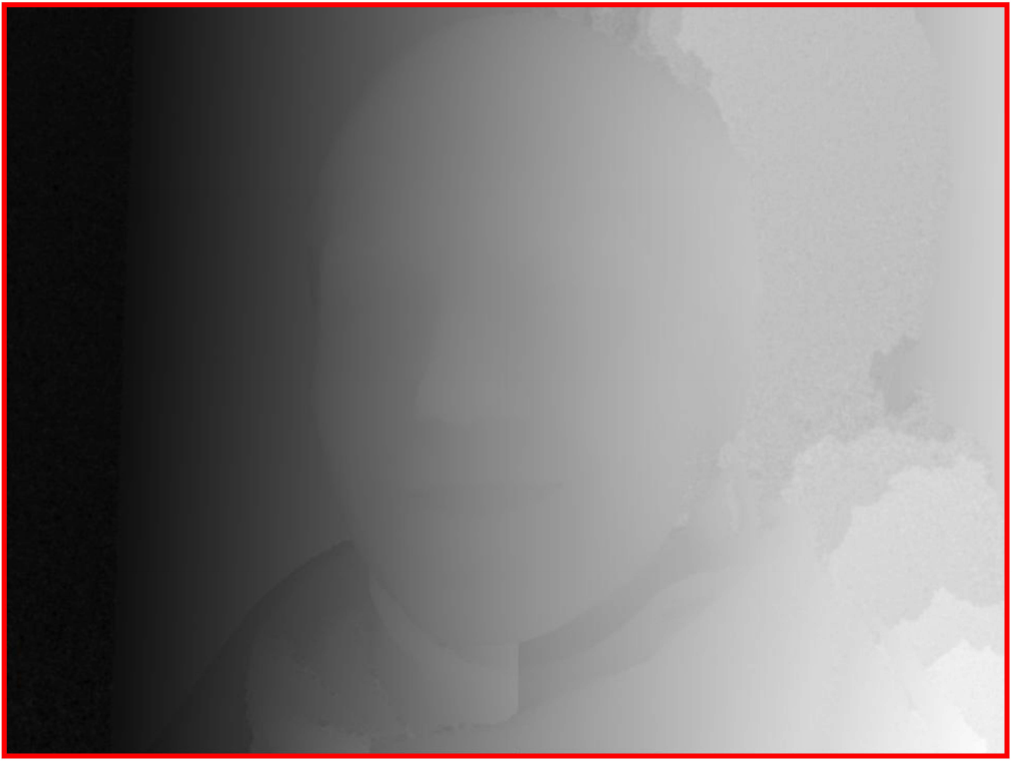} \\
        \includegraphics[width=0.32\textwidth]{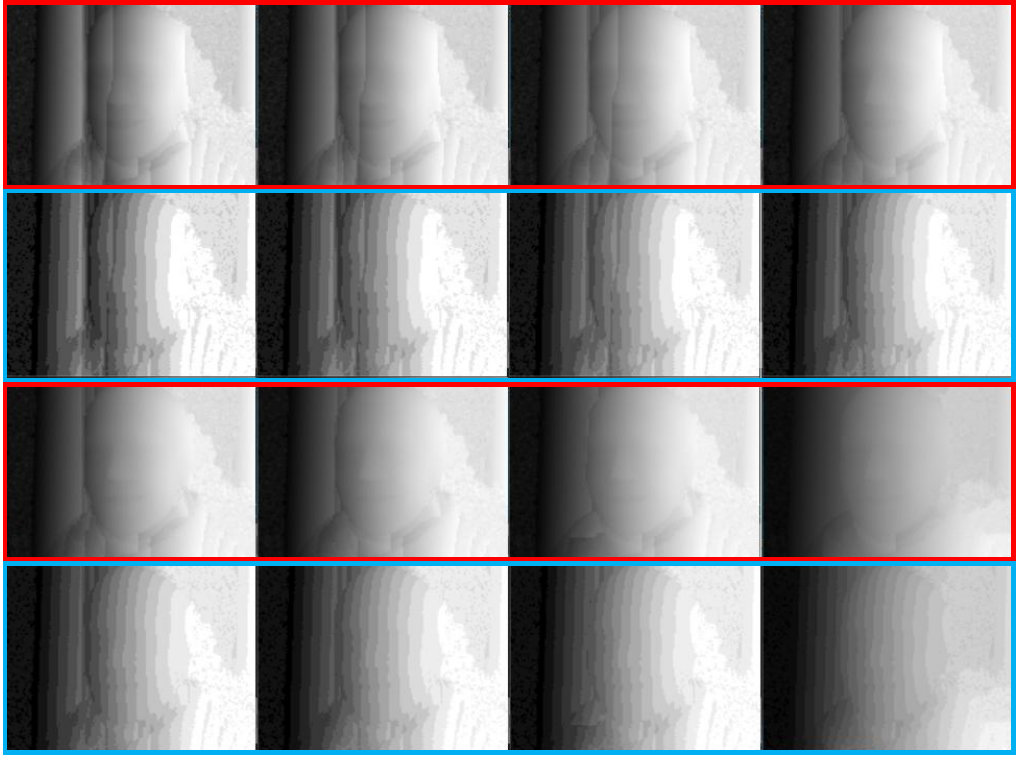} &
        \includegraphics[width=0.32\textwidth]{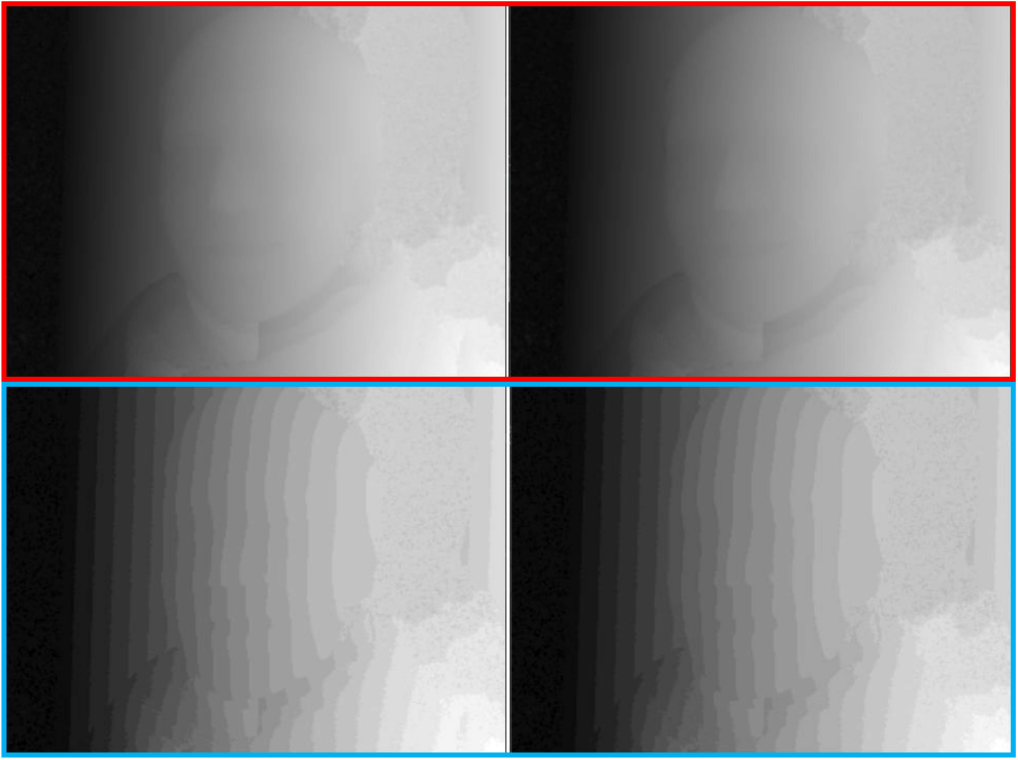} &
        \includegraphics[width=0.32\textwidth]{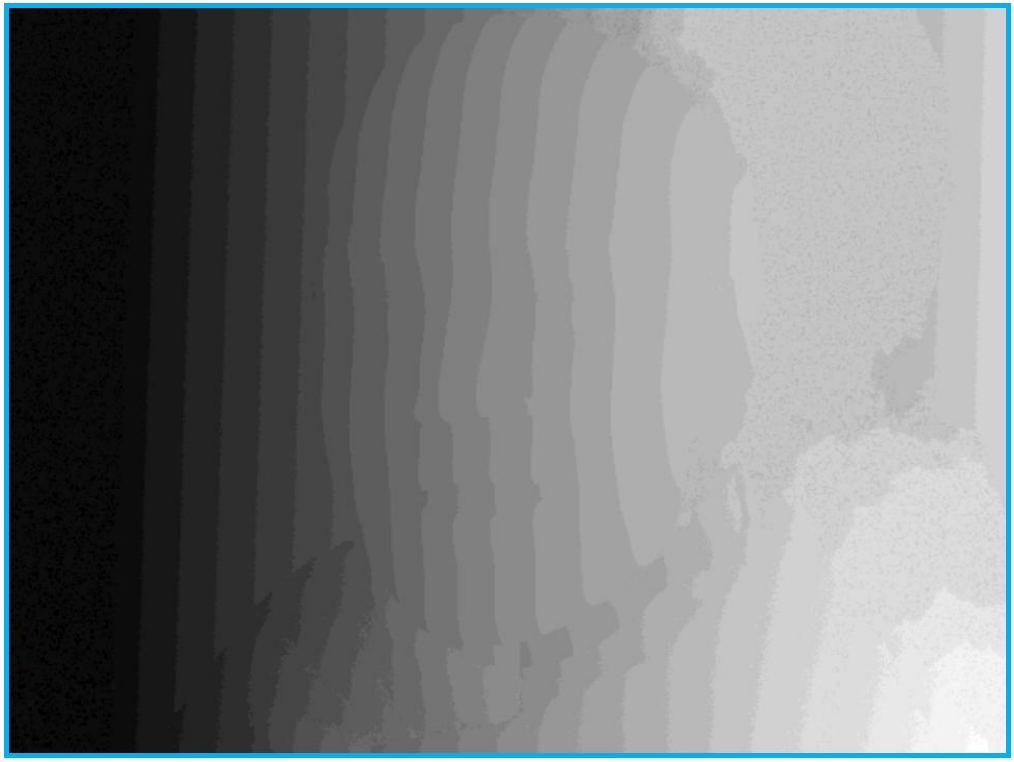} \\
        (a). \textbf{IDHGC - Level $0$} & (b). \textbf{IDHGC - Level $1$} & (c). \textbf{IDHGC - Level $2$} \\
    \end{tabular}
    \caption{ID Hierarchical GraphCut (\textbf{IDHGC}) phase unwrapping: Each column shows the temporary results at one level, sorted from left to right, and top to bottom. The top is the \textbf{Unwrapped Phase $\varPhi$}, and the bottom is the \textbf{Wrapped Count $k$}.}
    \label{fig:hierachical_graphcut_phaseunwrap}

    \vspace{5mm} 

    \begin{tabular}{cccc}
        \includegraphics[height=0.25\textwidth]{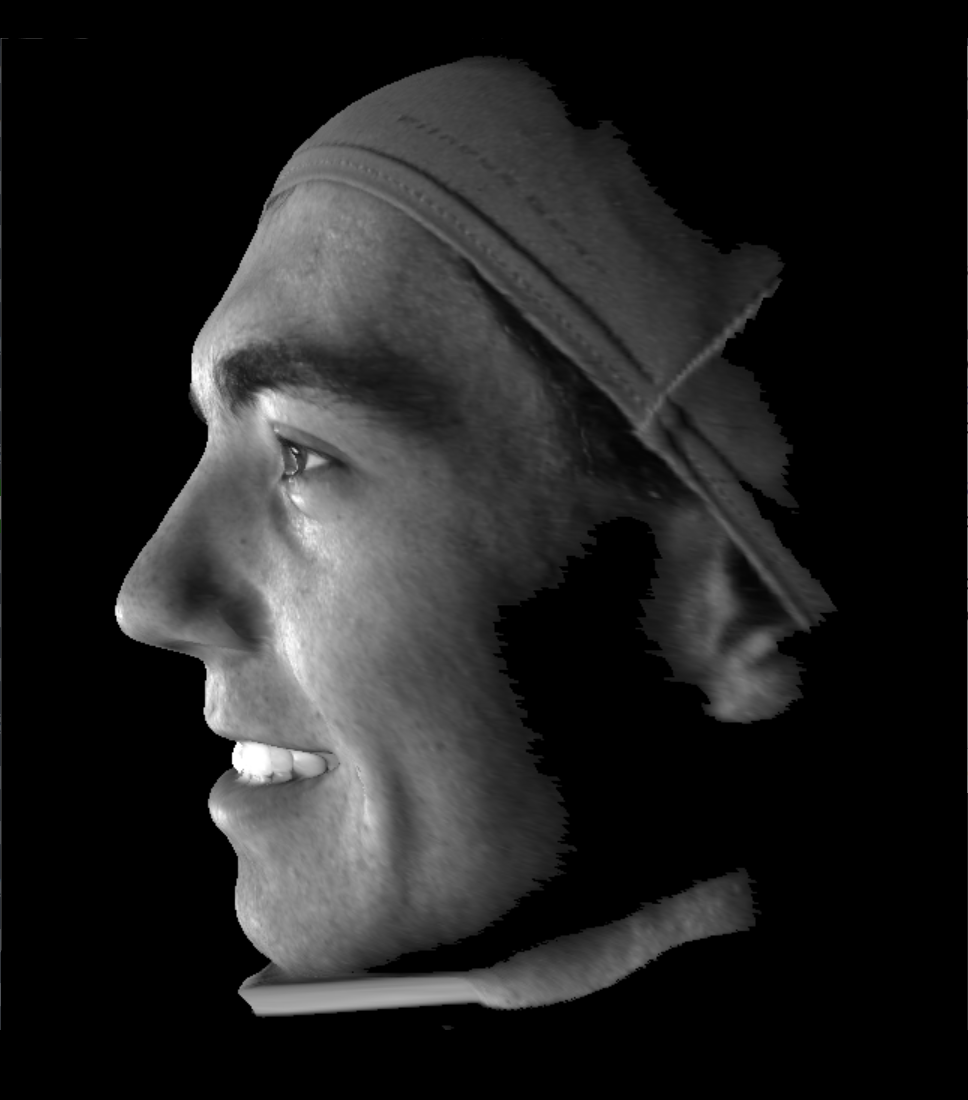} &
        \includegraphics[height=0.25\textwidth]{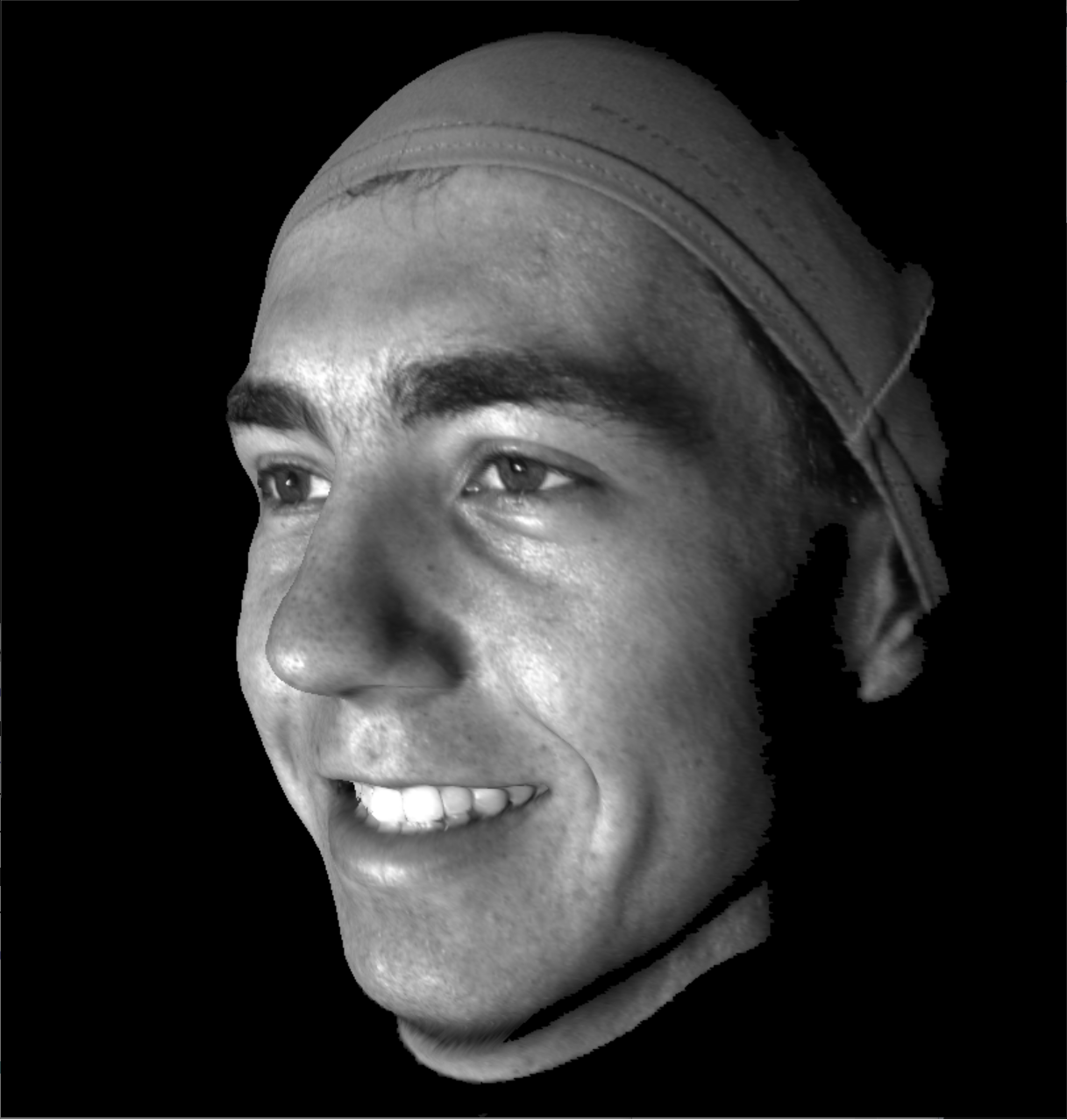} &
        \includegraphics[height=0.25\textwidth]{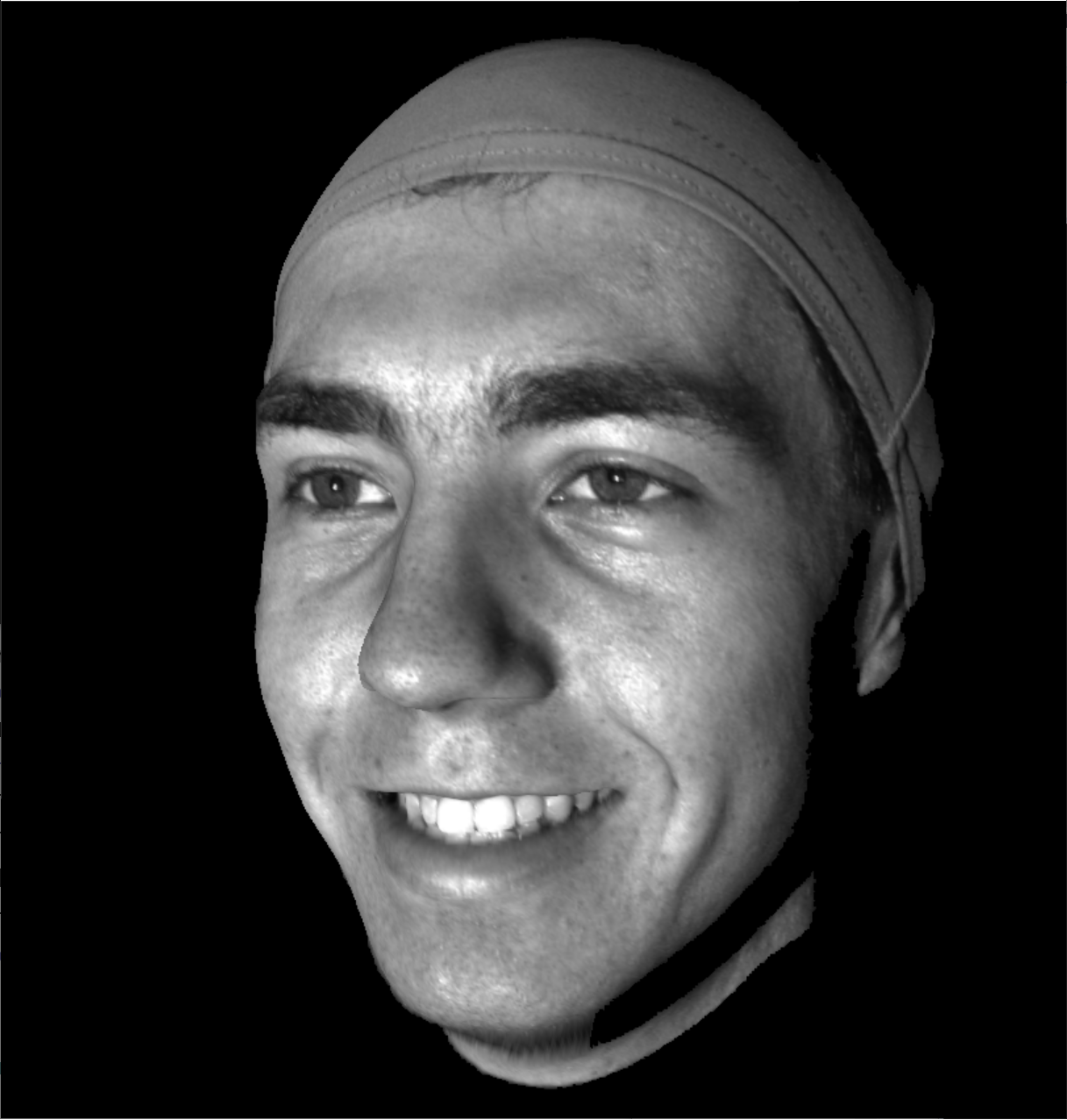} &
        \includegraphics[height=0.25\textwidth]{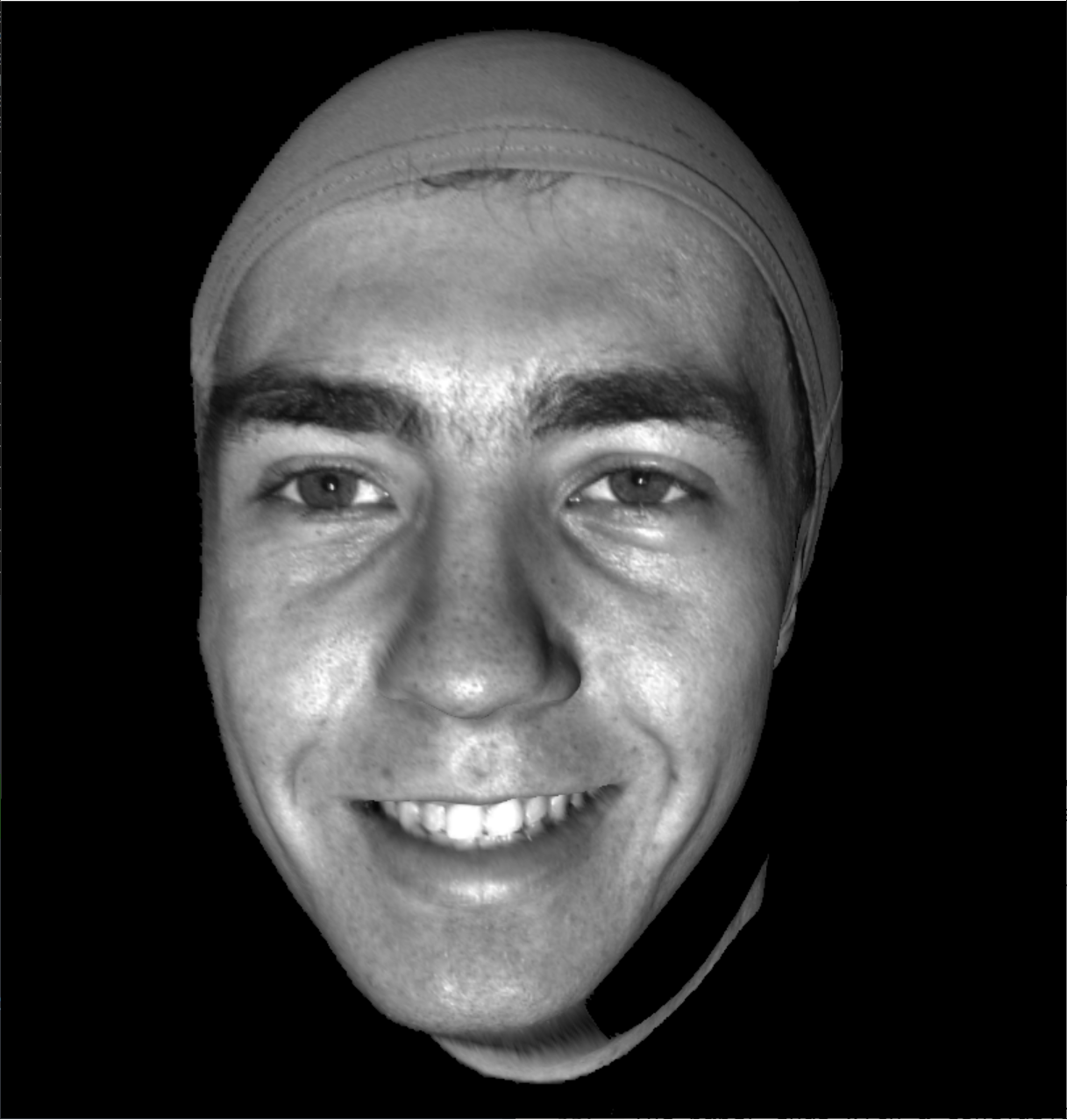} \\
    \end{tabular}
    \caption{Reconstructed 3D surface (Point Cloud) viewed from different camera angles, obtained using the unwrapped phase recovered by our method based on the wrapped phases data in \textbf{Fig.~\ref{fig:Andrew_phase_shifting}}. The ambient term \( A(p) \) is used as the texture.}
    \label{fig:Andrew_point_cloud}
\end{figure*}

\noindent{\textbf{Phase Shifting Structured Light Algorithm.}} In the active 3D vision field, the light source is a digital projector, 
which illuminates a physical object with structured light of special patterns. One of the most popular structured lights is the phase-shifting fringe pattern, the intensity of each pixel on the projector image is formulated as $L_k(u,v)=\frac{255}{2}(1+\cos(\frac{u}{\lambda}+\frac{2k\pi}{3}))$, $k=0,1,2$,
where $\lambda$ is the spatial wavelength. Digital cameras capture images of the object with fringe patterns. 
As shown in Fig.~\ref{fig:Andrew_phase_shifting}, the image formation can be modeled as $I_k(p)=A(p)+R(p)\cos(\Phi(p)+\frac{2k\pi}{3})$, 
where $A(p),R(p)$ are the \emph{ambient} component and the \emph{modulation} of the pixel $p$, $\Phi(p)$ is the \emph{absolute phase} of the pixel. Physically, each pixel $p$ on the camera image corresponds to a unique point on the object surface, which is lit by a ray of light emitted by the projector through a pixel $q$ on the projector image. The projector pixel $q$ is encoded by the absolute phase $\Phi(q)$, which equals the absolute phase $\Phi(p)$ at the camera pixel $p$. By the absolute phase $\Phi(p)$, we can find the corresponding pixel $q$ in the projector image. Then by the triangle principle, we can reconstruct the 3D point by intersecting the ray through the camera pixel $p$ and the ray through the projector pixel $q$. Therefore it is crucial to find the absolute phase $\Phi(p)$. However, from the raw fringe images, we can only directly recover the \emph{relative phase} (wrapped phase) by
\begin{equation}
\begin{split}
\varphi(p) &= \arctan \frac{\sqrt{3}(I_1(p)-I_2(p))}{2I_0(p)-I_1(p)-I_2(p)},\quad \\
    \Phi(p) &= \varphi(p) + 2\pi k(p), \quad k(p)\in \mathbb{Z}, 
    \label{eqn:absolute_phase}
\end{split}
\end{equation} where the integer function $k:\mathcal{P}\to\mathbb{Z}$ is called the \emph{phase count} function. The process of solving the phase count function is called \emph{phase unwrapping}. \textbf{Fig.~\ref{fig:Andrew_phase_shifting}} shows the phase shifting algorithm for a human facial surface. The top row shows the fringe images $I_k$'s, and the second row shows the ambient, modulation and the relative phase components.

We can find the phase count function by the energy optimization method. The energy is defined as 
\begin{equation}
    E(k) := \sum_{e_{pq}} E_{pq} = \sum_{e_{pq}} |\Phi(p)-\Phi(q)|^2,
\end{equation}
where $e_{pq}$ means the edge connecting the pixel $p$ and the pixel $q$, $\Phi(p)$ is the absolute phase of pixel $p$, and $E_{pq}$ is the edge energy defined as
\begin{equation}
    E_{pq}(k_p,k_q) = \psi_p(k_p) + \psi_q(k_q) + \psi_{pq}(k_p,k_q).
\end{equation}
Here, $\psi_p$ is the singleton energy and $\psi_{pq}$ is the edge energy, given by
\begin{equation}
    \psi_p(k_p) = k_p \sum_{e_{pq}} 4\pi (\varphi_p-\varphi_q),\quad \psi_{pq}(k_p,k_q) = 4\pi^2 (k_p-k_q)^2.
\end{equation}
This energy can be optimized using the GraphCut algorithm \cite{boykov2004experimental}. In order to improve the efficiency, we propose to use the hierarchical GraphCut algorithm, as shown in Fig.~\ref{fig:hierachical_graphcut_phaseunwrap}. Suppose $\varphi:\mathcal{P}\to\mathbb{R}$ is the input relative phase. We construct a pyramid representation of the original image, then unwrap the lower level relative phase and use the coarse level results as the initial guess for the higher level unwrapping. This hierarchical approach greatly improves the computational efficiency. Once the absolute phase $\Phi$ is obtained, the depth information can be recovered. \textbf{Fig.~\ref{fig:Andrew_point_cloud}} shows the point cloud reconstructed from the fringe images in \textbf{Fig.~\ref{fig:Andrew_phase_shifting}}. The ambient component in \textbf{Fig.~\ref{fig:Andrew_phase_shifting}(d)} is applied as the texture.

\noindent{\textbf{Framework of Invariance of Diffeomorphism. }} Given the fringe images $I_j$, $j=0,1,2$, we construct $n$ diffeomorphisms $g_i:\Omega\to \Omega$, $i=1,2,\dots,n$, using OT maps or conformal maps, and deform the images. We run the phase shifting algorithm on the deformed images $g_i(I_0),g_i(I_1),g_i(I_2)$ to compute the absolute phases $\Phi_i$'s and the phase counts $k_i$. We determine the final phase count function $k:\mathcal{P}\to \mathbb{Z}$ by the majority voting scheme: for each pixel $p\in \mathcal{P}$, we choose the majority of integers $\{k_1 \circ g_1^{-1}(p),\ k_2 \circ g_2^{-1}(p),\ \cdots,\ k_n \circ g_n^{-1}(p)\}
$ as the result of $k(p)$. In order to break possible ties, we choose $n$ to be odd. \textbf{Fig.\ref{fig:LV_ID_OT}} shows several diffeomorphisms constructed using OT maps. We select the region of interest (ROI) by drawing a circle $C(p,\sigma)$. The circle defines a Gaussian distribution $\mathcal{N}(p,\sigma^2)$, where the mean is the circle center $p$, and the standard deviation is the radius $\sigma$. Then we normalize the restriction of $\mathcal{N}(p,\sigma^2)$ on the image domain $\Omega$, and obtain the target measure $\nu$. We set the source measure $\mu$ to be the uniform distribution on $\Omega$ and use FFT-OT ~\cite{Lei2021ICCV} to compute the optimal transport map as shown in the second column. The relative phase $\varphi$ and the absolute phase $\Phi$ from the deformed images are shown in the third and fourth columns. Fig.~\ref{fig:LV_ID_CF} shows several diffeomorphisms constructed using the conformal maps, respectively. The rectangular fringe images are conformally mapped to the unit circle by a Riemann mapping using the Ricci flow algorithm in~\cite{Gu_JDG_2018}, then composed with different M\"obius transformations in Eqn.~(\ref{eqn:Mobius_transformation}).





\vspace{-3mm}
\section{Experiments and Results}

\subsection{Implementation Details}
We developed a real-time scanning system featuring a grayscale camera (Pike F-032), a color camera (Pike F-100), and a TI LightCrafter 4500 DMD projector, synchronized at 120 FPS using the phase shifting method. We implemented conformal mapping, optimal transport, and the ID Hierarchical GraphCut algorithms in C++, using \texttt{libfftw}~\cite{FFTW05} for Fast Fourier Transform (FFT), along with OpenGL and OpenCV for the user interface. Our framework operates in the image space and applies diffeomorphic transformations over \(N = 3\) region of interests (ROIs), which are chosen to be non-overlapping and spatially distinct. The image domain is conformally mapped to the unit disk using the surface Ricci flow algorithm~\cite{Zeng_TPAMI_2010}, and optimal transport maps are computed via the FFT-OT algorithm~\cite{Lei2021ICCV}. To evaluate computational efficiency, we synthesized a function \( f:\Omega \to \mathbb{R} \), computed the absolute phase \( \Phi = \frac{2\pi f}{\lambda} \), and derived the relative phase \( \varphi \) as \( \Phi \bmod 2\pi \). All experiments were conducted on an Intel I9-10900KF CPU and an NVIDIA GTX 1080 Ti GPU. For the \textbf{real experiment}, as recorded in Table~\ref{tab:efficiency_test}, we scan a human face from a single view to capture phase data, which are then used to compare our method against the baseline GraphCut \cite{graphcut2007}. In contrast, the \textbf{simulation setup} adds noise sampled from a Normal distribution to the double Gaussian phase map \textit{before wrapping} to simulate the acquisition process. Each method, including traditional phase unwrapping algorithms, is evaluated over three independent trials for their unwrapping $L^2$ error on the noise-corrupted phase map recorded in Table~\ref{tab:robustness_test}. 

\vspace{-4mm}

\subsection{Results}
\noindent \textbf{Performance and Speedups.} Both the GraphCut algorithm \cite{graphcut2007} and our proposed ID Hierarchical GraphCut were tested on $1024 \times 1024$ images with varying $\lambda$. The \textbf{ID Hierarchical GraphCut} demonstrated significant improvements in computational efficiency under the same number of periods of fringe projections. As shown in Table~\ref{tab:efficiency_test}, with increasing periods, our method achieved a \textbf{45.5$\times$ speedup} for \textbf{48 periods}, enabling the recovery of higher-resolution phases as the number of periods increases. Furthermore, quantitative results from Table~\ref{tab:robustness_test} demonstrate that the \textbf{ID Hierarchical GraphCut} not only achieves superior computational efficiency but also outperforms traditional methods, including GraphCut \cite{graphcut2007}, in terms of phase recovery accuracy by achieving the \textbf{lowest $L^2$ error}. These results highlight that our method is both highly efficient and capable of recovering accurate absolute phases.

\begin{figure}[t]
    \centering
    \begin{tabular}{cc}
    \includegraphics[width=0.195\textwidth]{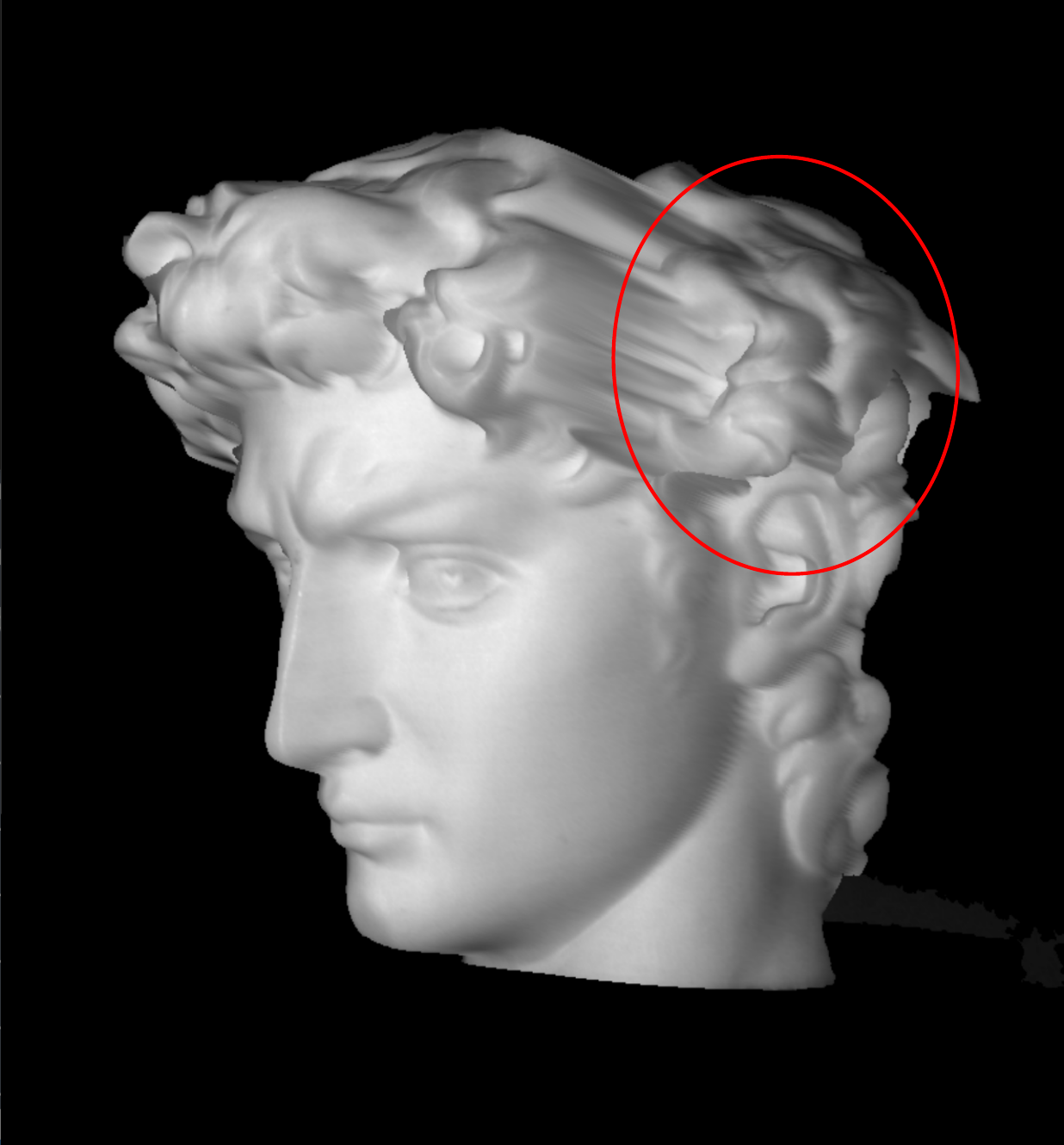} &
    \includegraphics[width=0.2\textwidth]{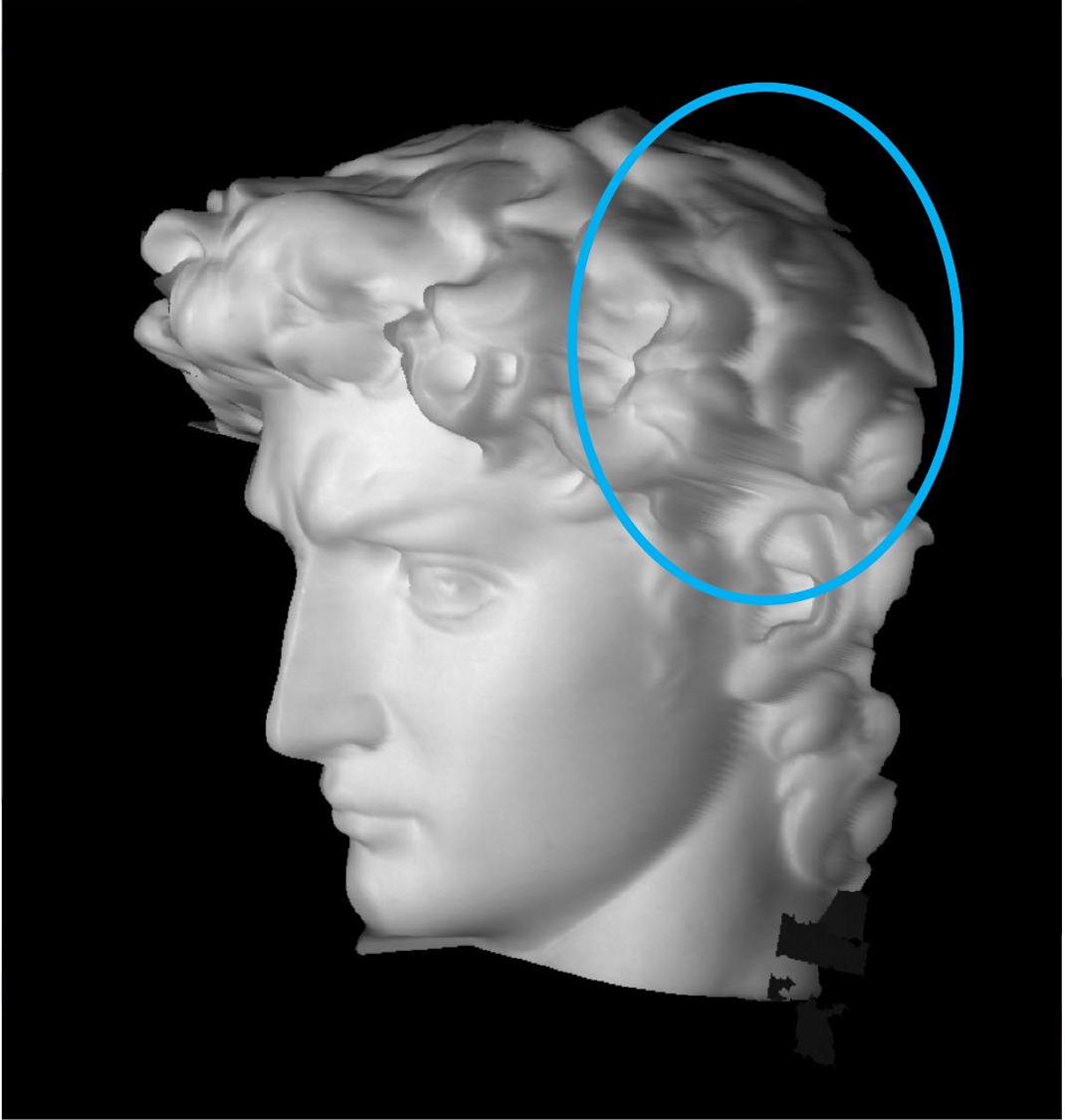} \\
    (a1). GraphCut \cite{graphcut2007} & (b1). \textbf{Ours} \\
    \includegraphics[width=0.2\textwidth]{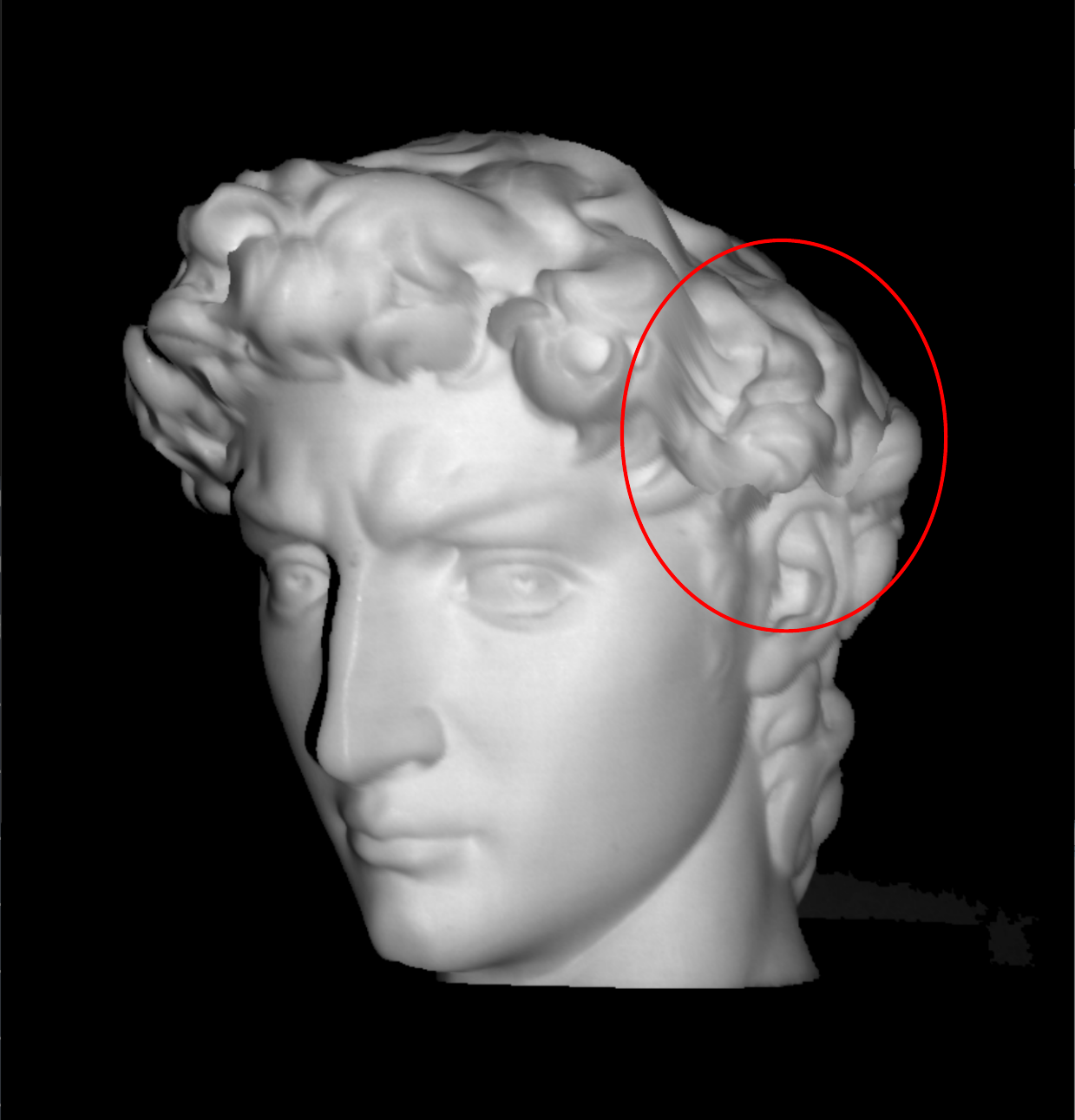} &
    \includegraphics[width=0.2\textwidth]{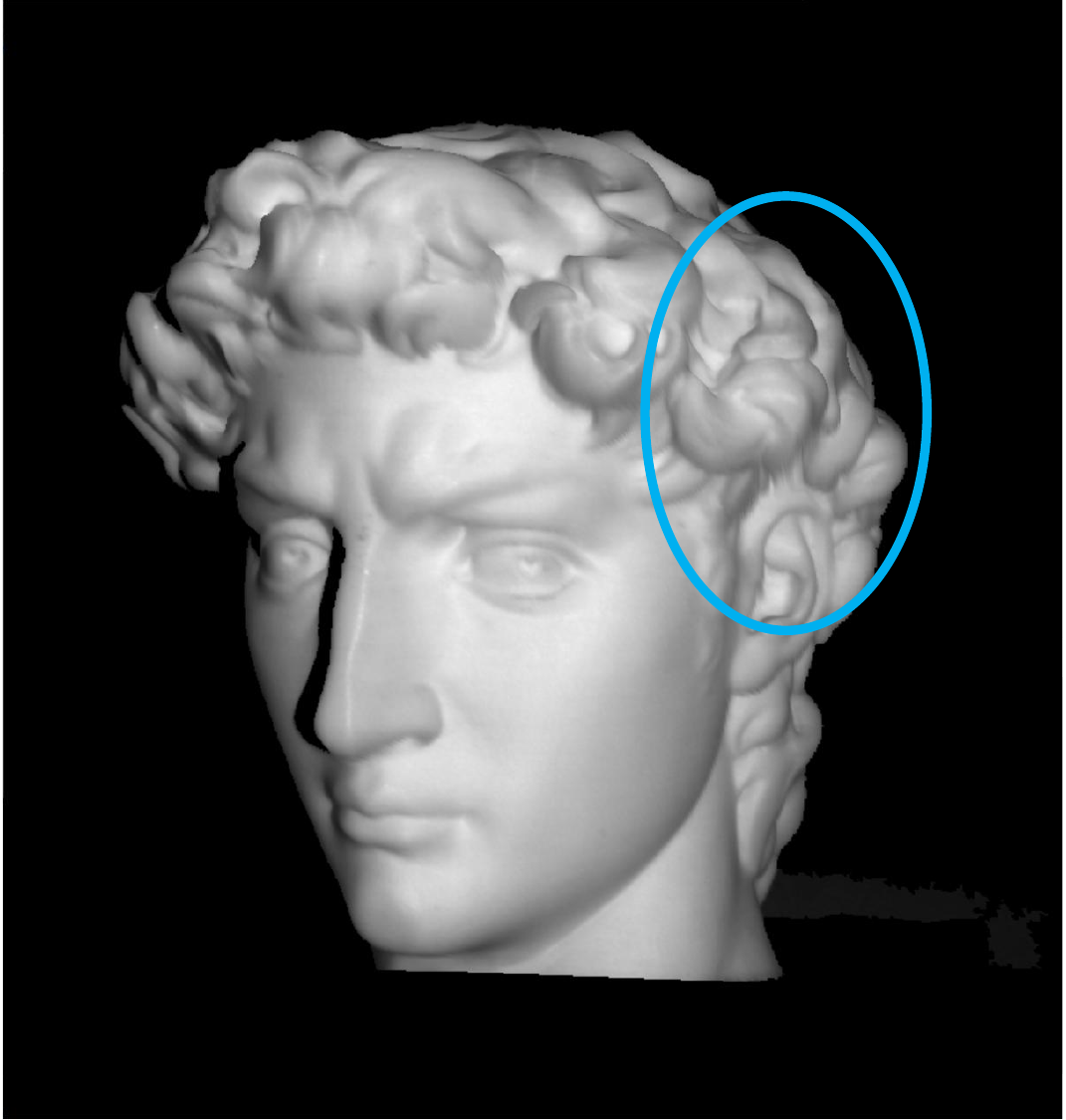} \\
    (a2). GraphCut \cite{graphcut2007} & (b2). \textbf{Ours} \\
    \end{tabular}
    \caption{Qualitative comparison between GraphCut \cite{graphcut2007} and our proposed \textbf{ID Hierarchical GraphCut}. The circled regions show sudden surface changes caused by unwrapping errors, which are corrected in our method.}
    \label{fig:IDGC_comparison}
\end{figure}

\noindent \textbf{3D Reconstructions.} Fig.~\ref{fig:LV_Point_cloud} and Fig.~\ref{fig:Andrew_point_cloud} illustrate that our ID Hierarchical GraphCut, combined with conformal mappings and optimal transport maps, accurately reconstructs high-quality 3D surfaces from single-view fringe projection. These qualitative results confirm the method’s ability to preserve fine surface details and reduce artifacts, consistent with the improved robustness and lower \( L^2 \) error demonstrated in both real and simulated experiments.

\noindent \textbf{Ablation Studies.} Existing phase unwrapping approaches have relied heavily on simulation for comparisons in terms of the \(L^p\)-norm error. In this study, we perform a real experiment that spans the full end-to-end pipeline: we scan the real David's head 3D model from multiple views to obtain the phase data, apply phase unwrapping algorithms to extract the unwrapped phase, and reconstruct the 3D surface for comparison against the baseline GraphCut method \cite{graphcut2007}. This enables a more realistic and practical evaluation of each method’s impact on the final 3D reconstruction quality. As demonstrated in Fig. ~\ref{fig:IDGC_comparison}, our method, leveraging diffeomorphisms, helps prevent sudden changes in the reconstructed 3D surface.
\begin{figure}[t]
    \centering
    \begin{tabular}{cc}
    \includegraphics[width=0.22\textwidth]{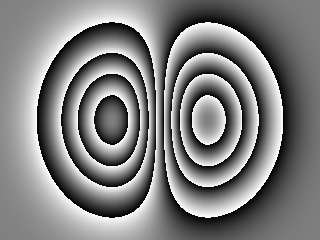} &
    \includegraphics[width=0.22\textwidth]{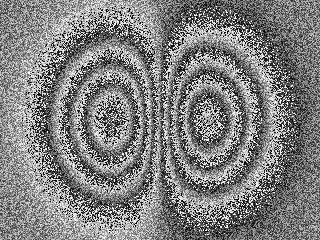} \\
    (a). Wrapped DG & (b). Noisy DG \\
    \includegraphics[width=0.22\textwidth]{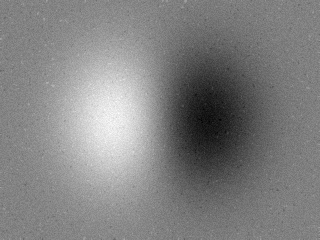} &
    \includegraphics[width=0.22\textwidth]{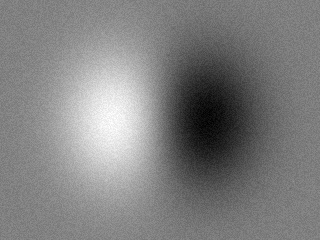} \\
    (c). GraphCut \cite{graphcut2007} & (d). \textbf{Ours} \\
    \end{tabular}
    \caption{Illustration of \textbf{Wrapped and Noise-Corrupted Double Gaussian}, along with unwrapped results using GraphCut \cite{graphcut2007} and our method. Our method achieves a lower $L^2$ error.}
    \label{fig:double_gaussian_comp}
\end{figure}

\begin{table}[h]
    \centering
    \caption{We report a quantitative comparison with GraphCut~\cite{graphcut2007} across different fringe periods. This comparison is conducted on \textbf{real experimental data} captured from 3D human face scans, using \(N = 3\) spatially distinct ROIs for unwrapping. Higher periods correspond to more fringe cycles, offering finer phase resolution at the cost of increased computational complexity. The table presents performance in terms of \textbf{absolute run time} and \textbf{speedup}.}
    \label{tab:efficiency_test}
    \footnotesize 
    \begin{tabular}{|c|c|c|c|c|c|}
    \hline
    \multirow{2}{*}{Periods} & \multicolumn{2}{c|}{GraphCut \cite{graphcut2007}} & \multicolumn{2}{c|}{ID Hierarchical GC (\textbf{Ours})} & \multirow{2}{*}{\textbf{Speedup}} \\ \cline{2-5}
     & Iter. & \textbf{Time (s)} & Iter. & \textbf{Time (s)} &  \\ \hline
    8  & 6   & 39.878  & 19   & \textbf{20.514}  & \textbf{1.944} \\ \hline
    16 & 12  & 83.526  & 24   & \textbf{11.216}  & \textbf{7.447} \\ \hline
    24 & 19  & 128.840 & 30   & \textbf{7.322}   & \textbf{17.555} \\ \hline
    32 & 23  & 200.719 & 35   & \textbf{7.275}   & \textbf{27.590} \\ \hline
    48 & 35  & 303.310 & 47   & \textbf{6.664}   & \textbf{45.515} \\ \hline
    \end{tabular}
\end{table}

\vspace{-3mm}
\begin{table}[htp]
    \centering
    \caption{Quantitative comparisons among traditional phase unwrapping algorithms, conducted as a \textbf{simulation} using Wrapped \textbf{Double Gaussian Phase Map} corrupted with noise as illustrated in Fig.~\ref{fig:double_gaussian_comp}. For each method, results are averaged over three independent trials, where in each trial, noise sampled from a Normal distribution is added to the phase map.}
    \label{tab:robustness_test}
    \footnotesize 
    \setlength{\tabcolsep}{6pt} 
    \renewcommand{\arraystretch}{1.3} 
    \begin{tabular}{|c|c|c|c|c|}
    \hline
    \textbf{Algorithm} & \multicolumn{4}{c|}{\textbf{Trials ($L^2$ Error)}} \\
    \cline{2-5}
     & \textbf{T1 ($L^2$)} & \textbf{T2 ($L^2$)} & \textbf{T3 ($L^2$)} & \textbf{Avg ($L^2$)} \\
    \hline
    FastUnwrap \cite{Ghiglia1998TwoDimensionalPU}  & 5.489  & 18.208 & 40.749 & 21.815 \\
    \hline
    MaskCut \cite{maskcut1996}      & 0.5209 & 0.2799 & 0.7504 & 0.5171 \\
    \hline
    QGPath  \cite{qualityguided1995}     & 0.5051 & 0.5127 & 0.5048 & 0.5075 \\
    \hline
    Goldstein \cite{Goldste1988}  & 0.2212 & 0.2213 & 0.2213 & 0.2213 \\
    \hline
    GraphCut \cite{graphcut2007}  & 0.1654 & 0.1755 & 0.1757 & 0.1722 \\
    \hline
    \parbox[c][0.75cm][c]{2.5cm}{\centering ID Hierarchical \\ GraphCut (\textbf{Ours})} & \textbf{0.1487} & \textbf{0.1510} & \textbf{0.1472} & \textbf{0.1483} \\
    \hline
    \end{tabular}
\end{table}

\section{Conclusion}
We demonstrated that our ID Hierarchical GraphCut method, leveraging diffeomorphism invariance via conformal and OT maps, achieves both high accuracy and efficiency, with a \textbf{45.5× speedup} and the \textbf{lowest $L^2$ loss} among existing methods. Ablation studies confirm its effectiveness for accurate 3D reconstruction of a full 3D object from multi-view scanned phase data. The method runs entirely on CPU, with future work aimed at GPU acceleration. This also shows the potential for processing 4D facial dynamics scanned data in real time.

\clearpage
\bibliographystyle{IEEEtran} 
\bibliography{main}  

\begin{IEEEbiography}[{\includegraphics[width=1in,height=1.25in,clip,keepaspectratio]{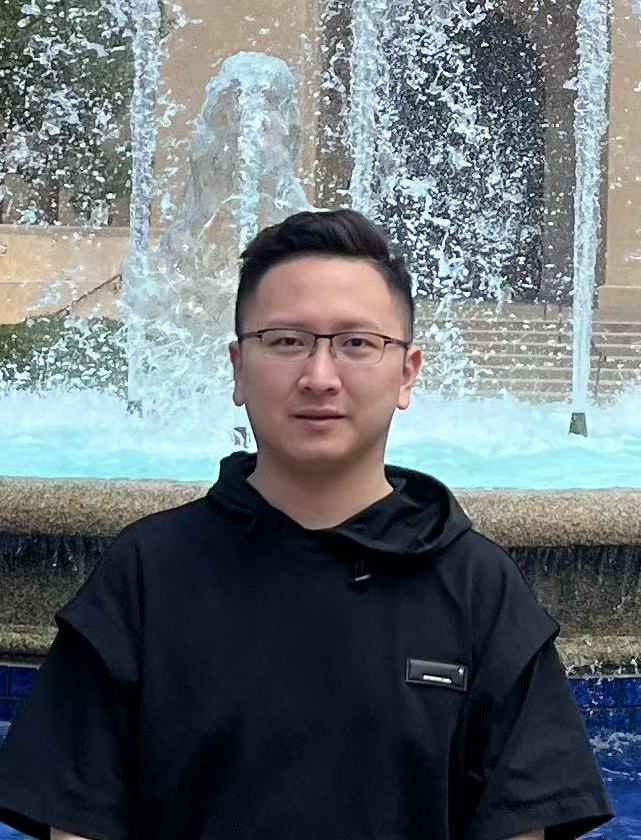}}]{Xiang Gao}
received his B.S. and M.S. degrees in Mechanical Engineering, and an M.S. degree in Applied Mathematics, all from Stony Brook University. He is currently a Ph.D. candidate in Computer Science at Stony Brook University advised by Prof. Xianfeng David Gu. His research interests include computer graphics, geometry processing, computer vision, and machine learning for 3D.
\end{IEEEbiography}

\begin{IEEEbiography}[{\includegraphics[width=1in,height=1.25in,clip,keepaspectratio]{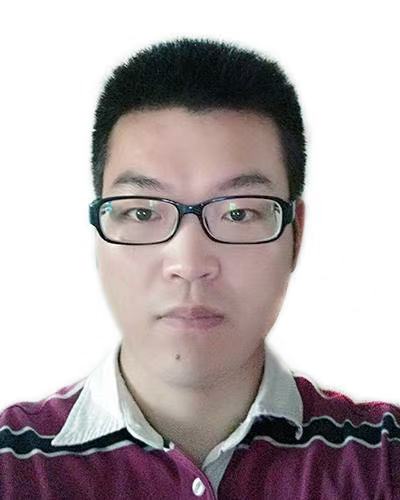}}]{Xinmu Wang}
received his B.S. degree from Shandong University and M.Eng. degree from Tsinghua University. He is currently a Ph.D. candidate at Stony Brook University advised by Prof. Xianfeng David Gu. His research interest includes computer vision, graphics, and deep learning.
\end{IEEEbiography}

\begin{IEEEbiography}[{\includegraphics[width=1in,height=1.25in,clip,keepaspectratio]{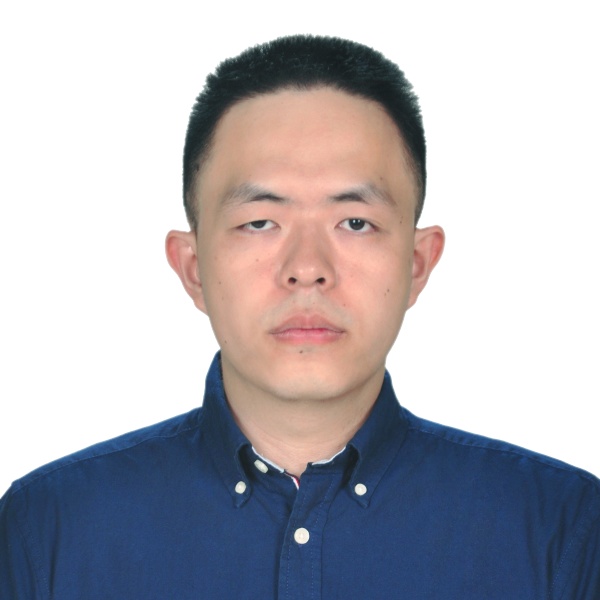}}]{Zhou Zhao}
received his B.S. degree from the University of Macau and his Ph.D. degree in Computer Science from Stony Brook University. He is currently a Research and Development Engineer at Ansys. His research interests include computational geometry, computer graphics, machine learning, and medical imaging.
\end{IEEEbiography}

\begin{IEEEbiography}[{\includegraphics[width=1in,height=1.25in,clip,keepaspectratio]{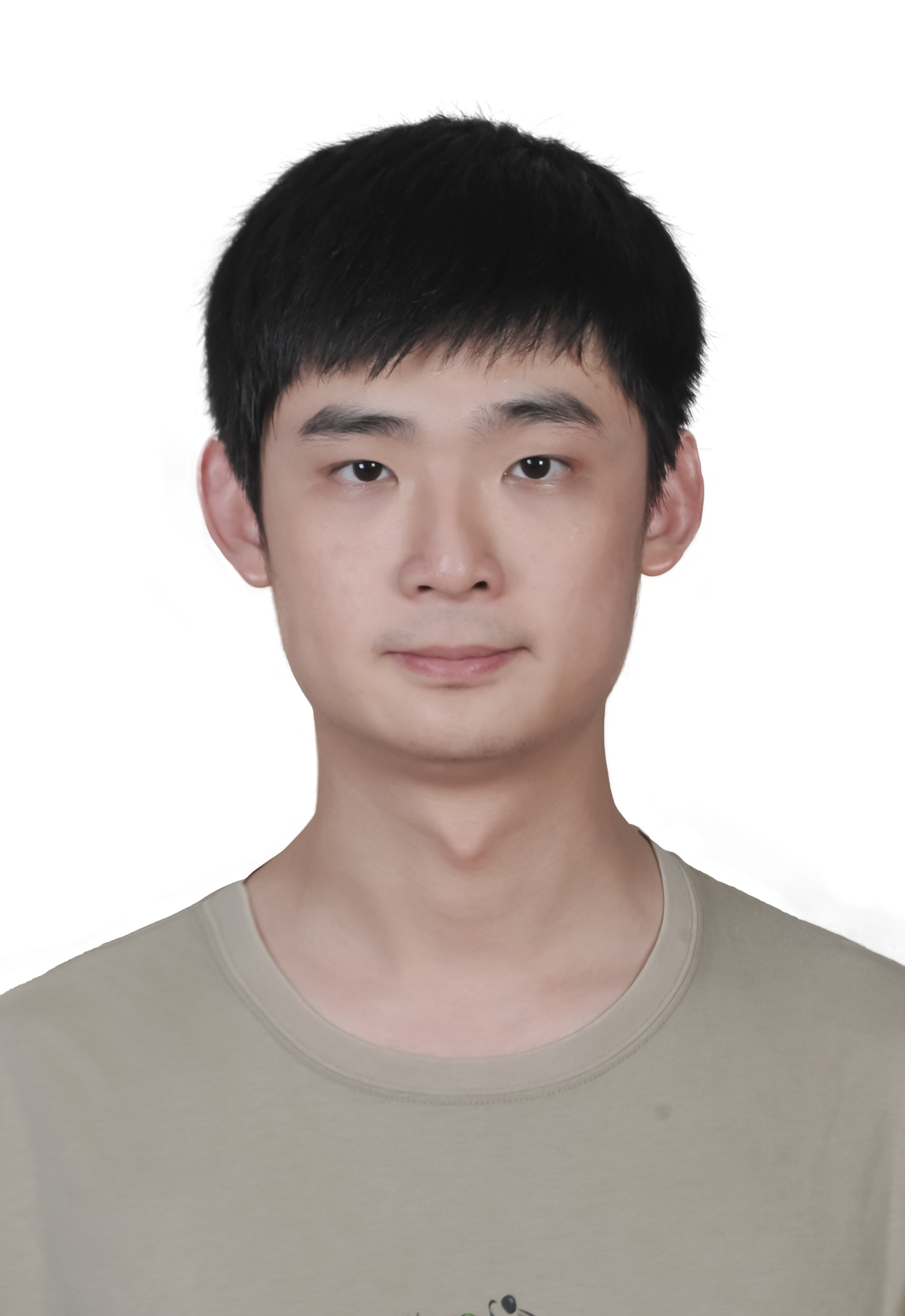}}]{Junqi Huang} is currently a Ph.D. candidate in the Department of Applied Mathematics and Statistics at Stony Brook University, advised by Prof. Xianfeng David Gu. He previously earned his Master’s degree in Mathematics from ETH Zurich. His research interests include optimal transport, mesh generation, and computational conformal geometry.
\end{IEEEbiography}

\begin{IEEEbiography}[{\includegraphics[width=1in,height=1.25in,clip,keepaspectratio]{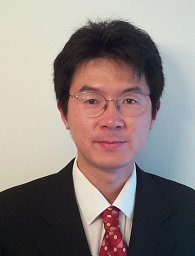}}]{Xianfeng David Gu}
received a B.S. degree from Tsinghua University and a Ph.D. degree from Harvard University. He is now a tenured professor
in the Department of Computer Science and Applied Mathematics at the State University of New
York at Stony Brook. He has won several awards,
such as the NSF CAREER Award of the USA,
the Chinese Overseas Outstanding Youth Award,
the Chinese Fields Medal, and the Chenxing Golden
Prize in Applied Mathematics. Professor Gu’s team
combines differential geometry, algebraic topology, Riemann surface theory, partial differential equations, and computer science to create a cross-disciplinary ``computational conformal
geometry'', which is widely utilized in computer graphics, computer vision, 3D geometric modeling and visualization, wireless sensor networks,
medical images, and other fields.
\end{IEEEbiography}

\end{document}